\documentclass{article}

\usepackage{booktabs}
\usepackage{amsmath} 
\usepackage{amssymb}
\usepackage{enumitem}
\usepackage{graphicx}
\usepackage{subcaption}
\usepackage{cuted}
\usepackage{caption}
\usepackage{stfloats}
\usepackage{wrapfig}
\usepackage{booktabs}

\usepackage{adjustbox}

\usepackage[final]{corl_2026} 

\makeatletter
\providecommand{\@makespecialcolbox}{\@make@specialcolbox}
\providecommand{\@makenormalcolbox}{\@make@normalcolbox}
\makeatother

\title{MorFiC: Fixing Value Miscalibration for Zero-Shot Quadruped Transfer}

\author{
Prakhar Mishra$^{1}$,
Amir Hossain Raj$^{2}$,
Xuesu Xiao$^{2}$,
and Dinesh Manocha$^{1}$\\
$^{1}$University of Maryland, College Park, MD, USA\\
$^{2}$George Mason University, Fairfax, VA, USA
}

\begin{document}
\maketitle






\vspace{-0.5em}
\begin{center}
\includegraphics[
    width=0.95\textwidth,
    height=0.22\textheight,
    keepaspectratio
]{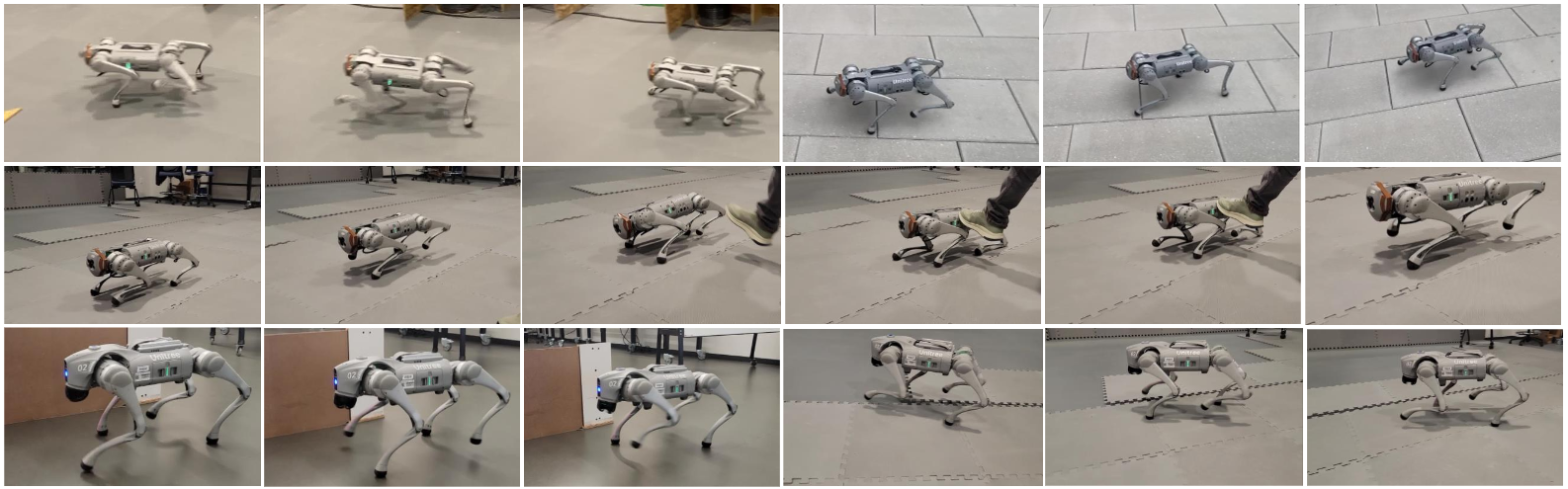}

\vspace{-0.6em}
\captionof{figure}{
We present \textbf{MorFiC}, a reinforcement learning framework trains on a single source quadruped and transfers zero-shot to multiple robots by improving the \emph{value function} via multiplicative conditioning. Real-world deployment on a Unitree Go1: MorFiC runs stably without task-specific fine-tuning.
\emph{Top two rows:} Go1 indoor/outdoor ($\approx 1.4$--$1.7$\,m/s).
\emph{Second row:} push-disturbance recovery on Go1.
\emph{Bottom row:} Zero-shot Go2 ($\approx 1.1$--$1.3$\,m/s) same policy weights.
}
\label{fig:qual_hardware}
\end{center}
\vspace{-1.3em}

\begin{abstract}
    Generalizing learned locomotion policies across quadrupedal robots with different morphologies remains a challenge. Policies trained on a single robot often fail when deployed on embodiments with different mass distributions, kinematics, joint limits, or actuation constraints, forcing per-robot retraining. Prior works have approached this primarily either by \emph{scaling} via training across real or generated embodiments or using \emph{large} architectures producing transferable policies which are both storage and compute heavy. We argue that both of these approaches circumvent a key failure mode in actor-critic learning: a shared value function tends to average incompatible value targets across embodiments, yielding miscalibrated advantages and fixing that helps transfer and reduce the compute cost. We present MorFiC, a reinforcement learning approach for zero-shot cross-morphology locomotion which fixes this issue using multiplicative critic conditioning on a morphology latent. Trained with a single source robot with morphology randomization in simulation, MorFiC achieves zero-shot transfer to total seven robots and reaching forward velocity competitive or surpassing scaling-based and morphology-conditioned PPO baselines for examples  $1.98$ m/s on AlienGo, where as additive PPO baselines remain below $0.65$ m/s. MorFiC achieves this using single training robot within 2 hours of training time compared to tens or hundreds of hours required for scaling baselines training on multiple embodiments. We diagnose critic calibration via three metric: Explained variance, advantage sign-flip rate and policy gradient cosine, confirming our critic conditioning as determinant for policy update direction. Finally, we demonstrate zero-shot deployment on Unitree Go1 and Go2 robots without fine-tuning.

\end{abstract}



\section{Introduction}
	
Reinforcement learning has allowed quadrupedal robots to achieve agile and robust locomotion across a wide range of speeds, terrains, and disturbances. Despite this progress, most of the learned locomotion policies remain tightly coupled to a single robot embodiment. Even within the same family of quadrupeds, policies trained on one platform often fail when deployed on another, substantially limiting policy reuse and increasing the deployment cost \citep{hwangbo2019learning, lee2020learning, rudin2022learning, pan2025roboduet}.

Generalizing locomotion policies across robot morphologies is challenging because relatively small changes in physical structure, such as mass distribution, joint limits, or actuation capacity, can significantly alter the underlying dynamics~\cite{yang2025multiloco}. Policies trained on a single robot tend to exploit morphology-specific regularities, which leads to brittle behavior when transferred to robots with different physical properties. Although domain randomization improves robustness to modeling errors and facilitates sim-to-real transfer \citep{tan2018sim, peng2018sim, tobin2017dr}, it does not explicitly address systematic shifts in dynamics caused by morphology variation. As a result, policies often degrade sharply when evaluated in robots whose morphology lies near or beyond the boundary of the training distribution.

A natural response to this limitation is to condition policies on explicit morphology descriptors, \emph{i.e.,} a vector of robot physical and control parameters (e.g., masses, joint limits, torque limits). The field has taken two main approaches to address this: (i) Scaling methods to train across multiple embodiments (real or generated) or (ii) using large architecture models and morphology randomization \citep{feng2023genloco, sferrazza2023body, shafiee2024many, ai2025towards, bohlinger2024onepolicy, shafiee2024manyquadrupeds, liu2025locoformer}. Training these approaches to achieve cross-morphology transfer comes with huge compute and storage cost and even then these morphology-agnostic policies might still generalize within in-family range of embodiments. However, in practice, reliable zero-shot transfer remains difficult, particularly for robots that differ substantially from those seen during training \cite{bohlinger2024onepolicy, feng2023genloco, ai2025towards, sferrazza2023body, liu2025locoformer}. 


We argue that this limitation arises not primarily from insufficient policy expressiveness but from miscalibrated value estimation (i.e., the model’s predicted long‑term return for a given state, action and morphology) across morphologies and in both of the current approach, a shared failure mode has been underexplored. We argue that the value function, not only the policy, should also generalize, in standard actor-critic methods such as PPO \citep{schulman2017ppo}, a single critic is trained to approximate expected returns in all training environments. When multiple morphologies are present, the same state can correspond to fundamentally different future outcomes depending on the robot’s physical properties. A shared critic is therefore forced to average incompatible value targets, inducing morphology-dependent bias in the value function. This bias propagates directly to the advantage estimates used for policy updates, destabilizing learning and disproportionately harming performance on edge-case morphologies \citep{henderson2018deep}.

We propose MorFiC which addresses this critic miscalibration via multiplicative conditioning using morphology latent. So our key findings are: (i) A diagnosis: We identify critic miscalibration as the poor cross morphology transfer, present in both scaling and large architecture based approaches. (ii) A method: MorFiC a multiplicative critic conditioning trained from single source robot and transfers to seven total quadrupeds using roughly two hours of GPU training significantly less that the current methods taking (40-300 hours) (iii) Evidence: critic diagnostic for the PPO and external baselines, a controlled actor only experiment with Monte carlo return, show why additive conditioning fails and zero-shot transfer  on Go1 and Go2 robots.


\section{Related Work}


\textbf{Learning-Based Quadrupedal Locomotion: }Reinforcement learning has enabled quadrupedal robots to achieve agile and robust locomotion, including high-speed running and disturbance recovery, both in simulation and on hardware \cite{rudin2022learning, lee2020learning, margolis2024rapid}. These successes are largely driven by large-scale simulation and careful reward design, but typically assume a fixed robot embodiment during training. As a result, the learned policies are highly robot-specific and often fail when deployed on robots with different morphologies. In contrast, MorFiC explicitly targets cross-morphology generalization by addressing value-function errors that arise when training across varying robot dynamics.

\textbf{Morphology-Conditioned and Multi-Robot Policies:} Several recent works explore generalizing locomotion policies across robot morphologies by conditioning on physical parameters or learned embodiment representations. URMA \cite{bohlinger2024one}, Embodiment Scale \cite{ai2025towards}, GenLoco \cite{feng2023genloco}, Body Transformer \cite{sferrazza2023body}, and ManyQuadrupeds \cite{shafiee2024many} demonstrate that a single policy can control multiple quadrupedal robots when provided with morphology information. However, these approaches rely on shared critics and assume that value functions generalize across morphologies.


\textbf{Value Learning, \& Training Stability :} Value-function accuracy is critical for stable actor-critic learning, yet its role in cross-morphology generalization has received limited attention in locomotion research. Previous work has shown that value misestimation can destabilize deep reinforcement learning \cite{henderson2018deep}, and recent studies highlight interference issues in shared critics in multiple tasks or environments \cite{liu2023multitask}. MorFiC addresses this issue directly by identifying morphology-induced value bias and mitigating it through morphology-aware critic modulation, leading to more stable advantage estimates and improved policy optimization. 


	
\section{Problem Setup \& Motivation}
 \label{sec:problem}

 We consider quadrupedal locomotion under morphology variation as a parameterized control problem. Each robot embodiment is identified by a vector of morphology parameters $m \in \mathcal{M}$ encoding physical and control properties (for example, mass distribution, joint limits, actuator limits). Training is performed on a morphology distribution $p_{\mathrm{train}}(m)$, while evaluation is performed on a holdout morphology $m^\star$ without additional training, fine-tuning, or online adaptation. 
 
 
 


\textbf{Formulation:}  Let locomotion across embodiments be modeled as a parameterized Markov Decision Process: $\mathcal{M}(m) = \langle \mathcal{S}, \mathcal{A}, P_m, R_m, \gamma \rangle$,
with state space $\mathcal{S}$, action space $\mathcal{A}$, morphology-dependent transition function $P_m : \mathcal{S} \times \mathcal{A} \to \mathcal{S}$ and reward function $R_m : \mathcal{S} \times \mathcal{A} \to \mathbb{R}$ and discount factor $\gamma \in (0,1)$. This formulation is closely related to contextual and hidden-parameter MDPs that model families of environments with shared structure but varying dynamics \cite{doshi2016hidden}. 





\textbf{Advantage Function:} Policies are trained using advantage-based actor--critic methods, via PPO optimization \cite{schulman2017ppo}. So, the advantage estimate: $\hat{A}_t = \hat{G}_t - V_\phi(s_t)$ becomes important, where $\hat{G}_t$ denote an empirical return estimate, $V_\phi$ is  learned value function. In practice, generalized advantage estimation is widely used to compute $\hat{A}_t$ with a favorable bias--variance tradeoff \cite{schulman2016gae}. Because the advantage weights the policy gradient, the systematic value error directly alters the direction and magnitude of policy updates \cite{schulman2017ppo}.






\textbf{Critic Miscalibration:} In multi-morph legged locomotion, the true value is morphology-dependent, $V^{\pi}(s_t,m)$~\cite{feng2023genloco}, 
but a shared critic with additive (concatenation)  or scaling-based conditioning tends to learn 
the average value
$\bar{V}_{\phi}(s_t) \approx 
\mathbb{E}_{m \sim p_{\mathrm{train}}}
\left[V^{\pi}(s_t,m)\right]$.
Since identical $(s_t,a_t)$ inputs can have different returns under different morphologies, 
a shared critic averages these incompatible targets, yielding morphology-dependent bias
$b_m(s_t)=\bar{V}_{\phi}(s_t)-V^{\pi}(s_t,m)$~\cite{yu2020gradient,shafiee2024many}. 
This bias propagates to the advantage:
$\hat{A}_t(m) \approx A_m^{\pi}(s_t,a_t)-b_m(s_t)$ which drives the PPO update, this produces inconsistent gradients across morphologies and unstable zero-shot transfer on OOD robots. 
This motivates us to fix the critic to represent morphology-specific returns within a single shared network.

\begin{figure*}[t]
\centering
\includegraphics[width=\textwidth]{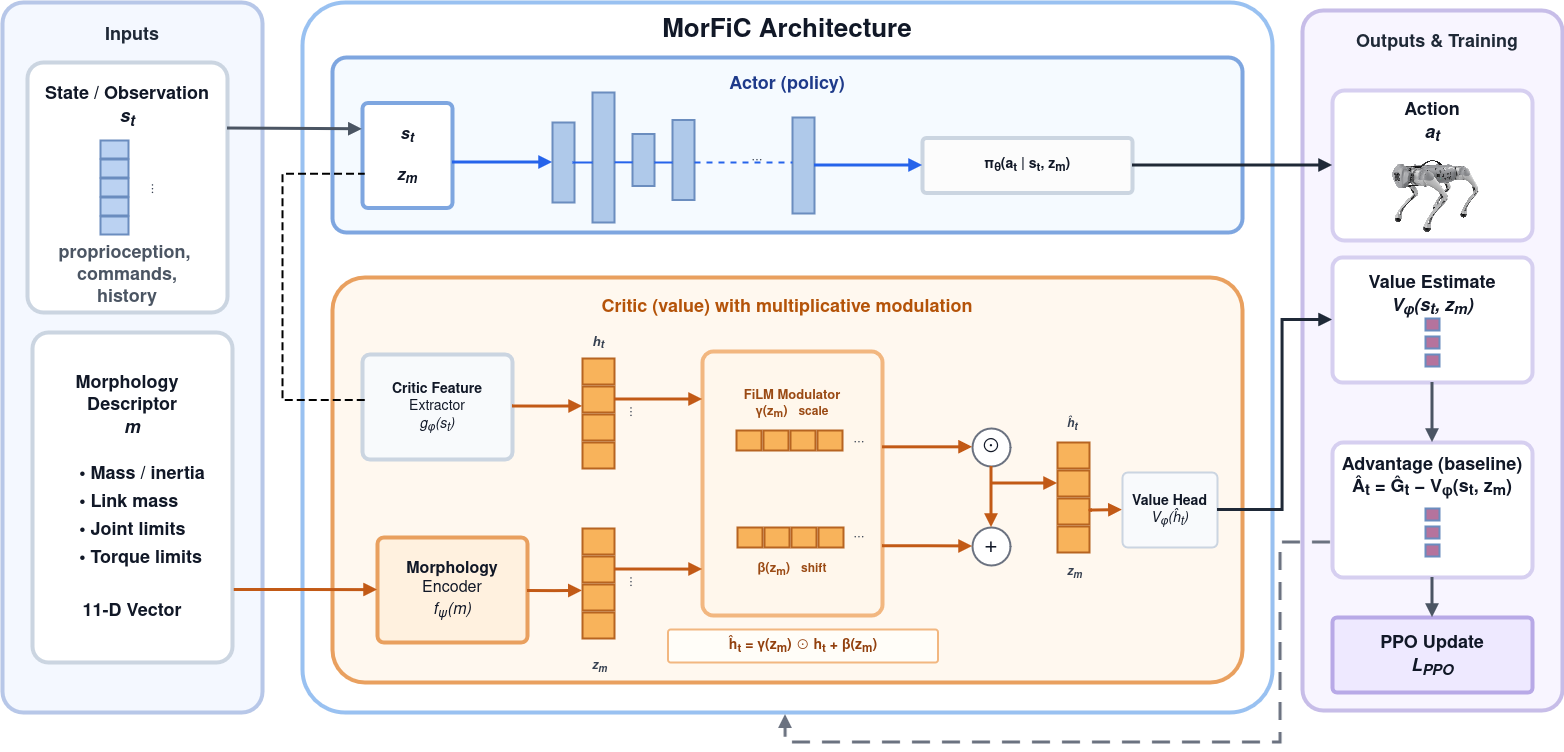}
\vspace{-0.7cm}
\caption{MorFiC Architecture. A morphology descriptor $m \sim p(m)$ is sampled, encoded via a morphology latent $z = f(m)$; and used to modulate the crtic via multiplicative conditioning; actor is also conditioned via same $z$.}
\label{fig:one_wide_image}
\end{figure*}

\section{MorFiC: Morphology Conditioned Critic Modulation}
\label{sec:morfic}

MorFiC is a morphology-aware actor--critic framework for zero-shot cross-morphology locomotion. The starting point is the failure mode in Section~\ref{sec:problem} critic miscalibration: MorFiC fixes this by conditioning the critic. This conditioning mechanism is derived in the rest of this section by (i) explicitly sampling morphology parameters and embedding, and (ii) addressing Additive Vs. Multiplicative principle (iii) critic Modulation Variants (iv) morphology-aware advantage.


\textbf{Design Requirement :}  A morphology-aware critic must satisfy two criteria: (i) representational sufficiency, so the critic must be able to represent distinct morphology landscape within the shared network (ii) PPO stability, the critic conditioning should not introduce noisy, or unstable value estimates during optimization, which destabilizes the  advantage updates $\hat{A}_t(m)$. In the rest of this section, we develop a morphology-conditioned critic modulation method which satisfies these two requirements while maintaining the efficiency of shared actor-critic network.


\textbf{Morphology Randomization \& Latent Encoder :} During training, each of the parallel environments samples a randomized morphology vector $m \sim p_{\mathrm{train}}(m)$ during reset, selected from a defined range (Appendix). The sampled normalized vector is applied to these environments in the simulator and kept fixed during the episode, so each robot acts as a single consistent embodiment. MorFiC maps $m$ to a learned latent representation $z_m = f_{\psi}(m)$, $z \in \mathbb{R}^{d_z},$ using a lightweight MLP. The same $z$ is provided to both the actor and the critic.

\textbf{Additive Vs. Multiplicative Principle:} To satisfy the design requirements, we need a critic to represent a morphology-indexed value functions. This gives us three possible choices: concatenating the morphology/other relevant input the critic, per-morphology heads and multiplicative conditioning. Per-morphology head can solve the value miscalibration challenge since it does not define critic for unseen morphology, it violates the zero-shot transfer and generalization argument. In concatenation, we simply append the $z_m$ vector to the input but this signal gets suppressed due to the large observation size. As Jayakumar et. al \citep{jayakumar2020multiplicative} contrast this additive limitation with multiplicative conditioning, for example $f(x,z) = W [x; z] + b$ where  $[x; z]$ represents concatenation of $x$ \& $z$, which can be further decomposed to \eqref{eq:add}, while the multiplicative interactions would have form: $f(x,z) = z^{\top} W x + z^{\top} U + Vx + b$, represented by \eqref{eq:mul1}, \eqref{eq:mul2}.


\vspace{-0.4em}
\begin{equation}
\begin{aligned}
f_{\mathrm{cat}}(x,z) &= W_x x + W_z z + b, \\
V_{\phi}^{\mathrm{cat}}(s_t,z_m) &= w_h^{\top} h_t + w_z^{\top} z_m + c,
\end{aligned}
\label{eq:add}
\end{equation}
\vspace{-0.3em}
\begin{equation}
\begin{aligned}
V_{\phi}^{\mathrm{mult}}(s_t,z_m)
&= z_m^{\top} W h_t + z_m^{\top} U + B h_t + c \\
\end{aligned}
\label{eq:mul1}
\end{equation}

\vspace{-1.5em}

\begin{equation}
\begin{aligned}
V_{\phi}^{\mathrm{mult}}(s_t,z_m) = A_m h_t + c_m ,
\qquad
A_m = z_m^{\top} W + B,
\qquad
c_m = z_m^{\top} U + c.
\end{aligned}
\label{eq:mul2}
\end{equation}

As its clear from the equation \eqref{eq:add}, there is no $h_t$ and $z_m$ interaction in additive conditioning, while multiplicative conditioning has $z_m^{\top} W h_t$, which is strictly missing from concatenation. So, Criterion (i), therefore, eliminates the additive conditioning, hence the critic must use multiplicative conditioning to represent the diverse value maps.

\textbf{Modulation Variants:} We have considered three modulation variants: FiLM, AdaLN and Hyper-network conditioning \citep{perez2018film,ha2016hypernetworks,ba2016layer}. These are the different approximation for the multiplicative principles rather than different conditioning. As Jayakumar et. al  describe gating like FiLM, AdaLN style modulation and hypernetworks as the variation of multiplicative interactions \citep{jayakumar2020multiplicative}.

A FiLM network modulates the critic features through morphology-conditioning via affine transformations and the $h_t$ from \eqref{eq:mul1} becomes:
\vspace{-0.2em}
\begin{equation}
\begin{aligned}
h_t^{\mathrm{FiLM}}
=
\gamma(z_m) \odot g_{\phi}(s_t) + \beta(z_m).
\end{aligned}
\label{eq:film}
\end{equation}
\vspace{-0.2em}
While AdaLN applies those morphology dependent affine transformation after normalizing the critic features and the $h_t$ becomes:
\vspace{-0.2em}
\begin{equation}
\begin{aligned}
h_t^{\mathrm{AdaLN}}
=
\gamma(z_m) \odot \mathrm{LN}\!\left(g_{\phi}(s_t)\right)
+
\beta(z_m).
\end{aligned}
\label{eq:adaln}
\end{equation}
\vspace{-0.2em}
So, both FiLM and AdaLN apply morphology dependent affine transformation, while AdaLN additionally controls feature via applying normalization. However, Hypernetworks directly generates the morphology-conditioned critic parameters using \eqref{eq:hyper}:
\vspace{-0.2em}
\begin{equation}
\begin{aligned}
V_{\phi}^{\mathrm{hyper}}(s_t,z_m) = A_m h_t + c_m,
\quad
(A_m,c_m) = H_{\eta}(z_m).
\end{aligned}
\label{eq:hyper}
\end{equation}
\vspace{-0.2em}
This allows the morphology to modulate the critic head itself allowing cross-feature mixing rather than just passive concatenation. Thus richer conditioning can improve morphology dependent generalization but might introduce higher variance value fitting. All the three variants satisfy the criterion (i) and the criterion (ii) PPO stability, is empirical and ablation section compare these. Since FiLM gives the best overall trade-off and we adopt it as the default instantiation of \textbf{MorFiC}: \textbf{Mor}phology-aware \textbf{Fi}LM \textbf{C}ritic. 



\textbf{Morphology-Aware Advantage Estimation:} MorFiC uses standard PPO~\cite{schulman2017ppo} without changing the sampling procedure, policy loss, or update rule. The only intervention is the value baseline used for advantage estimation $\hat{A}_t = \hat{G}_t - V_\phi(s_t,z_m)$, where $z_m$ is the morphology latent shared by the actor and critic. This replaces the morphology-averaged baseline $\bar{V}_\phi(s_t)$ with a morphology-aware critic, directly reducing the bias term $b_m(s_t)$ identified in Section~\ref{sec:problem}.

\section{Experimental Setup}

\textbf{Tasks and Robots :} We evaluate MorFiC on linear and angular command conditioned locomotion where policy tracks commanded forward/lateral linear and angular velocities. Commands are sampled via a curriculum at the start of the episode and gradually expanded during training and this mechanism is constant across all methods. And, An episode ends if the robot falls, collapses, or reaches time limit. We used Unitree Go2 robot as the base robot template with morphology randomization and evaluation is conducted on the source robot and the held-out robot morphologies like Go1, A1, Aliengo, MIT Mini Cheetah, ANYMal C robots.



\textbf{Evaluation Protocol :} Each parallel environment samples a morphology descriptor $m \sim p_{\mathrm{train}}(m)$ at the episode reset, applies the corresponding physical parameters to the source robot in the simulator, and keeps the descriptor same during the episode. Training is performed on a single source Go2 robot template with morphology randomization applied to that template. 


\textbf{Internal PPO baselines :} To isolate the effect of morphology conditioning, We have compared total six variants of PPO under identical training conditions, including MorFiC variants. These variants only differ in how the $z_m$ conditions the actor--critic architecture.

\begin{itemize}[leftmargin=1.2em,
                itemsep=0pt,
                parsep=0pt,
                partopsep=0pt,
                topsep=-1pt]
    \item \textbf{PPO:} No morphology conditioning, neither actor nor critic receives $z_m$.
    \item \textbf{Actor only:} Actor receives $z_m$; critic is unconditioned.
    \item \textbf{Actor + Critic} $z_m$ is concatenated to both actor and critic (additive conditioning).
    \item  \textbf{MorFiC-F:} Actor gets $z_m$, and critic is modulated via FiLM. 
    \item  \textbf{MorFiC-A:} Actor gets $z_m$, and critic is modulated via AdaLN. 
    \item  \textbf{MorFiC-H:} Actor gets $z_m$, and critic head is generated using a Hypernetwork. 
\end{itemize}


 \textbf{External baselines :}  In addition to the PPO variants, we compare MorFiC with four cross-morphology locomotion generalization methods namely: URMA \cite{bohlinger2024one}, GenLoco \cite{feng2023genloco}, BodyTransformer \cite{sferrazza2023body}, and Embodiment Scaling Laws \cite{ai2025towards}. These situate MorFiC relative to two dominant approaches: (i) Scaling the number of embodiments, or (ii) training on large models and neither of these approaches address value-function calibration.  For URMA, GenLoco, Embodiment Scaling Laws, we used the publicly available trained weights, but for BodyTransformer we trained their original available weights to 400M timestep to match MorFiC budget (see appendix).
 

\textbf{Metrics:} We utilized the following key metrics to measure the performance, survival rate and speed of MorFiC trained policy.  

\begin{itemize}[leftmargin=1.2em,
                itemsep=0pt,
                parsep=0pt,
                partopsep=0pt,
                topsep=-1pt]
    \item \textbf{Forward speed $\bar{v}$ (m/s):} We measured the forward velocity without tipping or falling on a given command velocity $v_{cmd}$.
    \item \textbf{Survival Rate $S$:} For simulation, we define it as the fraction of environment complete the given timestep. And in real-world setting, it's percentage of run, which were successful without any freezing or gait failure. Also, survival rate shall be better interpreted along with the speed because in some case the policy might have perfect survival rate without any feasible locomotion.
    \item \textbf{Explained Variance (EV) :} EV measures how well does the critic predict returns and is give by:  $\mathrm{EV}=1-\frac{\mathrm{Var}(G_t - V_{\phi}(s_t))}{\mathrm{Var}(G_t)}$.
\end{itemize}

\begin{figure*}[t]
    \centering
    \includegraphics[width=0.46\textwidth]{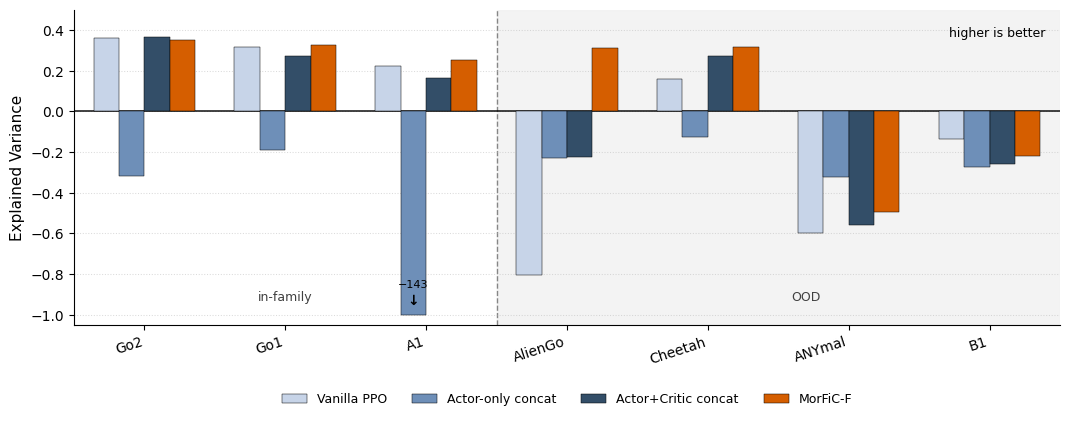}
    \hfill
    \includegraphics[width=0.52\textwidth]{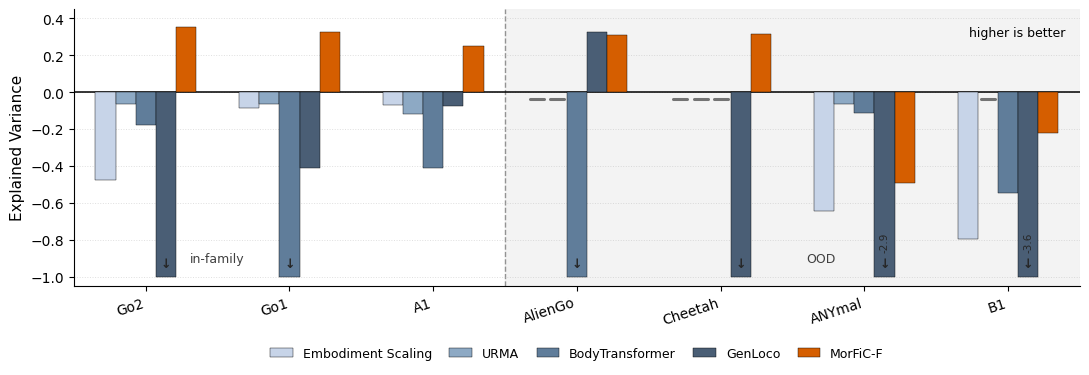}
    \caption{
    Critic calibration under morphology shift. Left: EV for internal PPO variants. Right: external baselines. Higher EV or closer to 1 is better.
    }
    \label{fig:critic_ev_wide}
\end{figure*}



\begin{figure*}[!b]
    \centering
    \includegraphics[
        width=0.46\textwidth,
        height=0.15\textheight,
        keepaspectratio
    ]{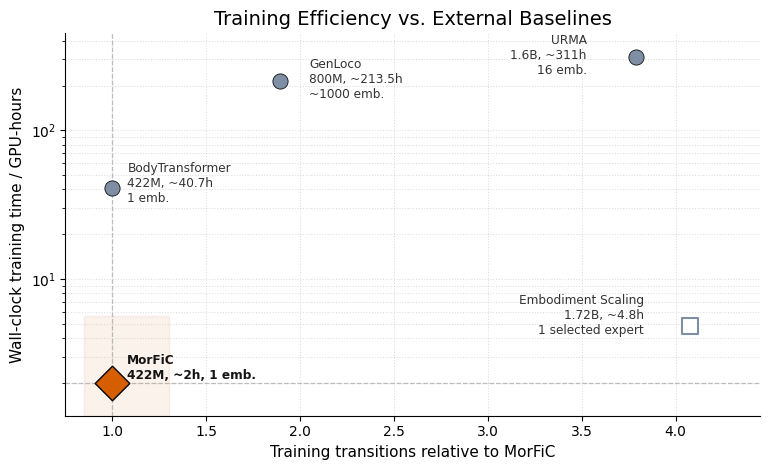}
    \hfill
    \includegraphics[
        width=0.52\textwidth,
        height=0.15\textheight,
        keepaspectratio
    ]{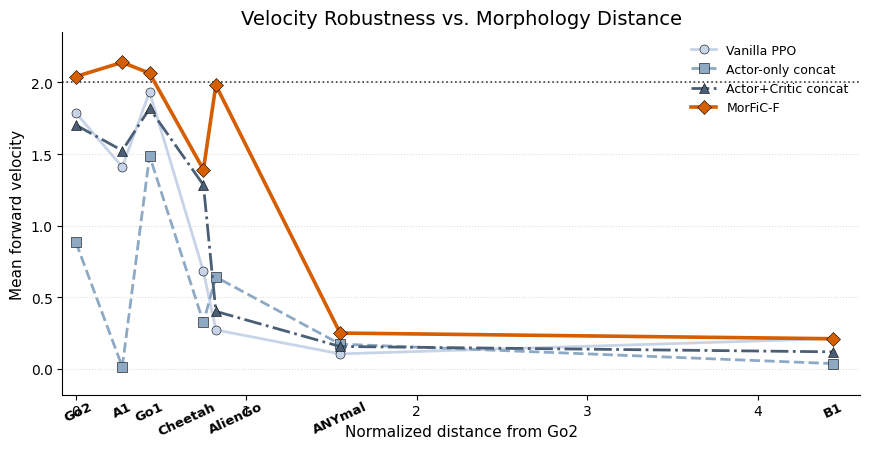}
    \vspace{-0.8em}
    \caption{
    Left: Compute cost (in hours) and total training timesteps used. Right: Performance w.r.t. normalized morphology distance. MorFiC performs better than both external and PPO baselines.
    }
    \label{fig:critic_ev_wide}
    \vspace{-1.0em}
\end{figure*}

\section{Results}

\textbf{Zero-shot cross morphology transfer: } Table~\ref{tab:zero_shot_transfer} compares MorFiC against three PPO variants for zero-shot transfer from Go2 training morphology to target robots. MorFiC achieves highest forward velocity on every target robot and highest gains come on OOD robots: for example on Aliengo it reaches $1.98$ m/s while rest of the additive PPO variants remain below $0.64$ m/s. B1 is the exception: all variant fail to have a proper locomotion gait.



\textbf{Critic Diagnostic:} We evaluated the critic miscalibration via explained variance (Fig. 3). Additive conditioning fails to calibrate as morphology shifts towards the heavier and out of distribution robots (Fig. 3). Multiplicative conditioning maintains calibration better and succeed on OOD robots like Cheetah and Aliengo. The Actor-only was most miscalibrated, indicating leaving critic unconditioned makes the overall return noisy and not good for transfer. 

Across external baselines, Most critic remain miscalibrated even when the transfer succeed, partly because successful targets fall within their in-family training distribution. So, these large model or embodiment scaling method do not solve the value miscalibration but rather circumvent via scaling, while MorFiC achieves competitive OOD results only by single source robot training and fixing the critic.


\noindent
\begin{minipage}[t]{0.48\linewidth}
\textbf{Causality analysis.}
To fully isolate the critic's effect, we freeze the MorFiC's actor policy and collected roll-out data on the target OOD Morphologies. We then varied the four PPO variant's critic to calculate the advantage sign flip rate against Monte-Carlo returns and gradient cosine values induced by the PPO variants' respective critics and report in  Table~\ref{tab:causal_adv}. MorFiC obtains the lowest flip rate and a high cosine value, sustaining the earlier argument that critic modulation is important for  policy update during morphology shift. 
\end{minipage}
\hfill
\begin{minipage}[t]{0.48\linewidth}
\centering
\footnotesize
\setlength{\tabcolsep}{3pt}
\renewcommand{\arraystretch}{0.9}
\captionof{table}{\textbf{Fixed-rollout diagnostic.} Sign Flip against wrong sign; Cos.: gradient cosine vs. MC.}
\label{tab:causal_adv}
\begin{tabular}{lcccc}
\toprule
 & \multicolumn{2}{c}{AlienGo} & \multicolumn{2}{c}{Cheetah}\\
\cmidrule(lr){2-3}\cmidrule(lr){4-5}
Src. & Flip & Cos. & Flip & Cos.\\
\midrule
MC      & 0.000 & 1.000  & 0.000 & 1.000\\
MorFiC  & \textbf{0.255} & \textbf{0.366}  & \textbf{0.209} & \textbf{0.679}\\
Concat  & 0.329 & 0.004  & 0.236 & 0.575\\
Act-only& 0.318 & 0.342  & 0.265 & 0.644\\
PPO     & 0.322 & -0.113 & 0.253 & 0.264\\
\bottomrule
\end{tabular}
\end{minipage}
\vspace{0.5em}




\textbf{Comparison to other Baselines.} Table~\ref{tab:cross_embodiment_baselines} compares MorFiC with the four current state of art baselines: URMA \cite{bohlinger2024one}, GenLoco \cite{feng2023genloco}, BodyTransformer \cite{sferrazza2023body}, and Embodiment Scaling Laws \cite{ai2025towards}.
MorFiC achieves the highest forward velocity on five of seven target robots. URMA despite strong on in-family distribution, still fails on AlienGo, B1 , Cheetah (with no valid forward motion). Embodiment Scaling also barely sustains motion on AlienGo, Cheetah but has surprising $2.86$ m/s on B1 robot. GenLoco remains consistent across six robots ($\leq 0.98$ m/s everywhere except on B1). While BodyTransformer remains weak on nearly all target robots. 


External baselines are computationally and storage expensive (Fig. 4), for example URMA roughly takes 311 hours, BodyTransfomrmer 40 hours, GenLoco 213 hours, while single expert policy of Embodiment Scale take 4.7 hours (1.7B) but training the full generalization would exceed more than 60 days and roughly 5TB to store those checkpoints. In contrast, we can train MorFiC in less than 2 hours with very little storage requirement.

\begin{table}[t]
\centering
\footnotesize
\setlength{\tabcolsep}{8.0pt}
\renewcommand{\arraystretch}{1.08}
\caption{
\textbf{Zero-shot cross-morphology transfer.}
Training is on Go2. Each cell reports $\bar{v}_x \pm \sigma / S$, where $\bar{v}_x$ is forward speed in m/s and $S$ is survival rate. Bold denotes the best speed and survival rate. We run all the PPO variants with ${v}^{cmd}_x = 2.0$ m/s for all 7 robots. All the experiments run for $500$ time-steps on total 64 environments.}
\label{tab:zero_shot_transfer}
\resizebox{\columnwidth}{!}{
\begin{tabular}{lcccc}
\toprule
\textbf{Target}
& \begin{tabular}{c}\textbf{PPO}\\[-1pt] $\bar{v}_x \pm \sigma / S$\end{tabular}
& \begin{tabular}{c}\textbf{Actor-only}\\[-1pt] $\bar{v}_x \pm \sigma / S$\end{tabular}
& \begin{tabular}{c}\textbf{Actor+Critic}\\[-1pt] $\bar{v}_x \pm \sigma / S$\end{tabular}
& \begin{tabular}{c}\textbf{MorFiC-F}\\[-1pt] $\bar{v}_x \pm \sigma / S$\end{tabular} \\
\midrule
Go2 
& $1.79{\pm}0.31/0.984$
& $0.88{\pm}0.23/0.265$
& $1.70{\pm}0.30/0.891$
& $\mathbf{2.04{\pm}0.37}/\mathbf{0.984}$ \\

Go1 
& $1.93{\pm}0.27/\mathbf{0.906}$
& $1.49{\pm}0.36/0.594$
& $1.82{\pm}0.32/0.859$
& $\mathbf{2.06{\pm}0.46}/0.547$ \\

A1 
& $1.41{\pm}0.42/0.406$
& $0.01{\pm}0.06/0.000$
& $1.52{\pm}0.27/0.422$
& $\mathbf{2.14{\pm}0.36}/\mathbf{0.516}$ \\

AlienGo 
& $0.27{\pm}0.07/0.000$
& $0.64{\pm}0.26/\mathbf{1.000}$
& $0.40{\pm}0.18/0.156$
& $\mathbf{1.98{\pm}0.51}/\mathbf{1.000}$ \\

B1 
& $0.21{\pm}0.06/\mathbf{1.000}$
& $0.04{\pm}0.07/\mathbf{1.000}$
& $0.12{\pm}0.07/\mathbf{1.000}$
& $\mathbf{0.22{\pm}0.08}/\mathbf{1.000}$ \\

Cheetah 
& $0.68{\pm}0.51/0.282$
& $0.32{\pm}0.49/\mathbf{1.000}$
& $1.28{\pm}0.58/0.382$
& $\mathbf{1.39{\pm}0.69}/0.422$ \\

ANYmal-C 
& $0.11{\pm}0.04/0.135$
& $0.17{\pm}0.08/0.469$
& $0.16{\pm}0.10/0.216$
& $\mathbf{0.25{\pm}0.21}/\mathbf{0.969}$ \\
\bottomrule
\end{tabular}
}
\end{table}


\begin{table}[t]
\centering
\small
\setlength{\tabcolsep}{1.0pt}
\renewcommand{\arraystretch}{1.05}
\caption{
\textbf{External baselines vs. MorFiC.}
 $\bar{v}_x \pm \sigma / S$ (m/s; survival). 
-- = no forward motion. 
$v_x^{\mathrm{cmd}} = 2.0$ m/s (GenLoco: no $v_x^{\mathrm{cmd}}$ input). 
Bold = best. }
\label{tab:cross_embodiment_baselines}
\resizebox{\columnwidth}{!}{
\begin{tabular}{lccccc}
\toprule
\textbf{Target}
& \begin{tabular}{c}\textbf{URMA}\\[-1pt] $\bar{v}_x{\pm}\sigma/S$\end{tabular}
& \begin{tabular}{c}\textbf{BodyTrans.}\\[-1pt] $\bar{v}_x{\pm}\sigma/S$\end{tabular}
& \begin{tabular}{c}\textbf{GenLoco}\\[-1pt] $\bar{v}_x{\pm}\sigma/S$\end{tabular}
& \begin{tabular}{c}\textbf{Embod.}\\[-1pt] $\bar{v}_x{\pm}\sigma/S$\end{tabular}
& \begin{tabular}{c}\textbf{MorFiC-F}\\[-1pt] $\bar{v}_x{\pm}\sigma/S$\end{tabular} \\
\midrule
Go2
& $1.84{\pm}0.26/\mathbf{1.000}$
& $0.30{\pm}0.21/0.656$
& $0.59{\pm}0.05/0.922$
& $1.23{\pm}0.06/0.906$
& $\mathbf{2.04{\pm}0.37}/0.984$ \\

Go1
& $1.93{\pm}0.28/\mathbf{1.000}$
& $0.21{\pm}0.13/\mathbf{1.000}$
& $0.98{\pm}0.01/\mathbf{1.000}$
& $1.94{\pm}0.09/0.953$
& $\mathbf{2.06{\pm}0.46}/0.547$ \\

A1
& $1.79{\pm}0.25/\mathbf{1.000}$
& $0.39{\pm}0.15/0.421$
& $0.92{\pm}0.01/\mathbf{1.000}$
& $1.86{\pm}0.08/0.906$
& $\mathbf{2.14{\pm}0.36}/0.516$ \\

AlienGo
& $-/-$
& $0.30{\pm}0.28/0.234$
& $0.96{\pm}0.03/\mathbf{1.000}$
& $0.41{\pm}0.08/0.000$
& $\mathbf{1.98{\pm}0.51}/\mathbf{1.000}$ \\

B1
& $-/-$
& $0.04{\pm}0.11/0.796$
& $0.00/0.000$
& $\mathbf{2.86{\pm}0.11}/0.969$
& $0.21{\pm}0.08/\mathbf{1.000}$ \\

Cheetah
& $-/-$
& $-/\mathbf{1.000}$
& $0.73{\pm}0.06/0.969$
& $0.06{\pm}0.09/0.000$
& $\mathbf{1.39{\pm}0.69}/0.422$ \\

ANYmal-C
& $\mathbf{1.80{\pm}0.28}/\mathbf{1.000}$
& $0.01{\pm}0.06/0.875$
& $0.71{\pm}0.01/\mathbf{1.000}$
& $0.44{\pm}0.15/0.656$
& $0.25{\pm}0.21/0.969$ \\
\bottomrule
\end{tabular}
}
\end{table}

\begin{table}[t]
\centering
\small
\setlength{\tabcolsep}{8.0pt}
\renewcommand{\arraystretch}{1.05}
\caption{
\textbf{Ablation Table.}
Each cell reports $\bar{v}_x \pm \sigma / S$, where $\bar{v}_x$ is forward speed in m/s and $S$ is survival rate. Alive time is omitted. Bold denotes the best speed per target robot.
}
\label{tab:critic_conditioning_ablation}
\resizebox{\columnwidth}{!}{
\begin{tabular}{lcccc}
\toprule
\textbf{Target}
& \begin{tabular}{c}\textbf{MorFiC-F}\\[-1pt] $\bar{v}_x{\pm}\sigma/S$\end{tabular}
& \begin{tabular}{c}\textbf{MorFiC-F(Scaled)}\\[-1pt] $\bar{v}_x{\pm}\sigma/S$\end{tabular}
& \begin{tabular}{c}\textbf{MorFiC-H}\\[-1pt] $\bar{v}_x{\pm}\sigma/S$\end{tabular}
& \begin{tabular}{c}\textbf{MorFiC-A}\\[-1pt] $\bar{v}_x{\pm}\sigma/S$\end{tabular} \\
\midrule
Go2
& $2.04{\pm}0.37/0.984$
& $2.03{\pm}0.26/0.875$
& $\mathbf{2.10{\pm}0.28}/0.953$
& $2.03{\pm}0.19/\mathbf{1.000}$ \\

Go1
& $\mathbf{2.06{\pm}0.46}/0.647$
& $1.98{\pm}0.34/0.823$
& $1.99{\pm}0.31/0.766$
& $1.99{\pm}0.35/\mathbf{0.853}$ \\

A1
& $\mathbf{2.14{\pm}0.36}/0.516$
& $2.04{\pm}0.24/0.531$
& $2.02{\pm}0.26/\mathbf{0.766}$
& $1.76{\pm}0.26/0.609$ \\

AlienGo
& $\mathbf{1.98{\pm}0.51}/\mathbf{1.000}$
& $1.78{\pm}0.45/\mathbf{1.000}$
& $1.75{\pm}0.46/\mathbf{1.000}$
& $1.79{\pm}0.52/\mathbf{1.000}$ \\

B1
& $\mathbf{0.21{\pm}0.08}/\mathbf{1.000}$
& $0.18{\pm}0.08/\mathbf{1.000}$
& $0.20{\pm}0.03/\mathbf{1.000}$
& $0.18{\pm}0.06/\mathbf{1.000}$ \\

Cheetah
& $1.39{\pm}0.69/0.422$
& $\mathbf{1.97{\pm}0.62}/0.359$
& $1.61{\pm}0.45/0.344$
& $1.57{\pm}0.56/\mathbf{0.703}$ \\

ANYmal
& $0.25{\pm}0.21/0.969$
& $0.24{\pm}0.06/\mathbf{1.000}$
& $\mathbf{0.46{\pm}0.19}/\mathbf{1.000}$
& $0.39{\pm}0.28/\mathbf{1.000}$ \\
\bottomrule
\end{tabular}
}
\end{table}


\section{Ablations}

MorFiC instantiates multiplicative conditioning with FiLM critic but to test whether the transfer affected by the specific choice or the broader multiplicative-conditioning principle. We evaluated three other multiplicative variants and report them in Table~\ref{tab:critic_conditioning_ablation}. Our results show that improvements are not FiLM specific, all other variants like hypernetworks, AdaLN and Scaled FiLM achieve similar transfer performance across held-out morphologies. FiLM leads on Go1, A1, AlienGo, the hypernetwork on ANYMal and Go2 and AdaLN was very high on survival rate and stability. We adopt FiLM as the default instantiation since it provided higher velocity on four robots and stable PPO updates, satisfying our design requirements in section III. 



\section{Real-Robot Results}



\textbf{Zero-shot Hardware Deployment.} We deploy MorFiC policy on two hardware platforms namely Go1 and Go2 using an onboard LCM stack for state estimation and actions. We deploy the final policy on robots without any finetuning, MorFiC reaches a nearly $\approx 1.4$--$1.7$\,m/s on Go1 robot over 10 trials with a survival rate of over 80\% while on Go2 robots it reaches roughly $\approx 1.1$--$1.3$\,m/s. We deployed the same go2 policy weights which we used for zero-shot evaluation in sim-transfer. 





\textbf{Failure Modes \& Analysis :} We have observed two categories of failures, System-level (generic) failure and Robot specific failures (Go2). The generic failures are mainly due to i) Latency and state-estimator drift, ii) Contact mismatch during runs (either foot slips or uneven gait). For robot specific failures, Go2 exhibits a reduced  speed and also a higher failure rate under our current deployment pipeline.

\section{Limitations}

MorFiC relies on morphology randomization during training and we found transfer depends heavily on the normalized distance, so Go2 trained robot transfer well to nearby quadrupeds well (Go1, A1, Cheetah) but fails on distant targets like B1 and Anymal. Since this distance in our case was heavily influenced by base mass and joint limits, even increasing those limits during randomization made policy conservative (Appendix). Training robot choice anchors the transfer direction, Go2 trained policy transfer well to mid-size robots but fails on heavier one; a heavier training robot would likely have inverted pattern. A natural extension to that is combine MorFiC to limited multi-source training rather than large set of robots in scaling methods. Finally we evaluate MorFic transfer on the quadrupeds locomotion tasks; extending the framework to bipeds and humanoids is left to future work.

\section{Conclusion} 
\label{sec:conclusion}

We have studied the zero-shot cross morphology problem and found that transfer failure is strongly driven by the critic miscalibration, leads to biased advantage, and ultimately leads to poor PPO updates. MorFiC addresses this by conditioning the critic function via multiplicative conditioning.  Based on our results, MorFic improves survival rate, speed, and OOD generalization better than any PPO variant or external baseline while remaining highly competitive in family robot transfer and transfers zero-shot to 2 real robots (Go1 \& Go2). 

\section*{Author Contributions}

\noindent\textbf{Prakhar Mishra:}
Conceptualization, methodology, implementation, experimental design, and analysis.

\noindent\textbf{Amir Hossain Raj:}
Experimental implementation, robot deployment, evaluation, and discussions.

\noindent\textbf{Xuesu Xiao and Dinesh Manocha:}
Supervision, technical guidance, and manuscript review.




\bibliography{example}  

\clearpage
\appendix





    
    

\section{Evaluation Metrics and Diagnostic Definitions}

\textbf{Normalized morphology distance.} To analyze how transfer performance degrades with morphology shift, we have defined the normalized distance $d \in [0,1]$ between target robot and training robot based on the 11-D morphology descriptor selected from Table I. We compute.
\begingroup
\setlength{\abovedisplayskip}{0pt}
\setlength{\belowdisplayskip}{0pt}
\setlength{\abovedisplayshortskip}{0pt}
\setlength{\belowdisplayshortskip}{0pt}

\begin{equation}
d(m_{\mathrm{target}}, m_{\mathrm{train}})
=
\frac{1}{\sqrt{11}}
\left\|
\tilde{m}_{\mathrm{target}} - \tilde{m}_{\mathrm{train}}
\right\|_2 .
\label{eq:morph_dist}
\end{equation}
\endgroup
This distance is used for transfer loss analysis and is not provided to the policy or critic during training (refer to Fig. 5).



\paragraph{Normalized Value Bias (NVB).} NVB measures whether the critic systematically overestimates or underestimates returns, where close to 0 equals low systematic bias and higher value indicate higher bias and is given by:

\begin{equation}
 \mathrm{NVB}_\phi = \frac{ \mathbb E_{(t,i)\in\mathcal I} \left[ V_\phi(s_{t,i},z_{m_i})-\hat G_{t,i} \right] }{ \sigma_G+\epsilon }. 
\label{eq:nvb}
\end{equation}

\paragraph{Normalized Value Error (NVE).} NVE measures the total critic prediction error, normalized by the return scale, lower value means better calibrated critic and is given by:

\begin{equation}
\mathrm{NVE}_\phi = \frac{ \sqrt{ \mathbb E_{(t,i)\in\mathcal I} \left[ \left( V_\phi(s_{t,i},z_{m_i})-\hat G_{t,i} \right)^2 \right] } }{ \sigma_G+\epsilon }.. 
\label{eq:nve}
\end{equation}

\paragraph{Advantage Sign Agreement (ASA).} ASA measures whether the critic-induced advantage has the same sign as the Monte-Carlo reference advantage, higher ASA is better and is given by:

\begin{equation}
\hat{A}^{\mathrm{MC}}_{t,i}
=
\hat{G}_{t,i}
-
\frac{1}{|\mathcal{I}_{\epsilon}|}
\sum_{(t',i')\in\mathcal{I}_{\epsilon}}
\hat{G}_{t',i'} .
\end{equation}

\begin{equation}
\hat{A}^{\phi}_{t,i}
=
\hat{G}_{t,i}
-
V_{\phi}(s_{t,i}, z_m).
\end{equation}

\begin{equation}
\mathrm{ASA}_{\phi}
=
\frac{1}{|\mathcal{I}_{\epsilon}|}
\sum_{(t,i)\in\mathcal{I}_{\epsilon}}
\mathbf{1}
\left[
\operatorname{sign}\!\left(\hat{A}^{\phi}_{t,i}\right)
=
\operatorname{sign}\!\left(\hat{A}^{\mathrm{MC}}_{t,i}\right)
\right].
\end{equation}

\paragraph{Sign Flip Rate.} Sign flip rate measures how often the critic-induced advantage has the opposite sign from the MC reference advantage and lower rate is better and is given by:

\begin{equation}
\mathrm{Flip}_{\phi}
=
\frac{1}{|\mathcal{I}_{\epsilon}|}
\sum_{(t,i)\in\mathcal{I}_{\epsilon}}
\mathbf{1}
\left[
\operatorname{sign}\!\left(\hat{A}^{\phi}_{t,i}\right)
\neq
\operatorname{sign}\!\left(\hat{A}^{\mathrm{MC}}_{t,i}\right)
\right].
\end{equation}

\begin{equation}
\mathrm{Flip}_{\phi}
=
1-\mathrm{ASA}_{\phi}.
\end{equation}

\paragraph{Gradient Cosine.} Gradient cosine measures whether the PPO actor update direction produced by the critic-induced advantage matches the actor update direction produced by the MC reference advantage, where close to 1 means critic gives almost the same actor update as MC return, so a value higher than 0 and closer to 1 is better and is given by:

\begin{equation}
g(\hat{A}^{\phi})
=
\nabla_{\theta}
\mathcal{L}_{\mathrm{actor}}
\left(\theta; \hat{A}^{\phi}\right),
\qquad
g(\hat{A}^{\mathrm{MC}})
=
\nabla_{\theta}
\mathcal{L}_{\mathrm{actor}}
\left(\theta; \hat{A}^{\mathrm{MC}}\right).
\end{equation}

\begin{equation}
\mathrm{CosGrad}_{\phi}
=
\frac{
g(\hat{A}^{\phi})^{\top} g(\hat{A}^{\mathrm{MC}})
}{
\left\|g(\hat{A}^{\phi})\right\|_2
\left\|g(\hat{A}^{\mathrm{MC}})\right\|_2
}.
\end{equation}

\paragraph{Validity of survival-based metrics.}
Survival rate alone is not sufficient to indicate successful locomotion under cross-morphology transfer. 
In several in \& out-of-distribution cases, a policy can avoid early termination while producing negligible forward motion, unstable stepping, collapsed postures, or physically invalid behaviors such as dragging or non-propulsive oscillations. And in some cases the policy do have valid motions while also producing behaviors like upside down leg movement, forward falling, disabled rear legs etc in some fraction of robot while the rate remains 1.00 or 100\%.
We therefore interpret survival jointly with achieved forward velocity, tracking ratio, and rollout validity. In the result tables, high survival with near-zero forward velocity is treated as a failure to transfer rather than successful locomotion. 

This distinction is especially important for distant morphologies such as B1 and ANYmal-C, where policies may remain non-terminated but fail to generate a valid gait. High survival with near-zero forward velocity is not considered successful transfer; survival is interpreted jointly with achieved speed and rollout validity.

\newcommand{\hi}[1]{\textbf{#1}$\uparrow$}
\newcommand{\lo}[1]{\textbf{#1}$\downarrow$}

\begin{table*}[t]
\centering
\scriptsize
\setlength{\tabcolsep}{3.2pt}
\caption{
\textbf{Morphology randomization parameters.}
Each row reports one dimension of the 11-D morphology vector used by MorFiC.
Values outside the training support are shown in bold with $\uparrow$ / $\downarrow$.
}
\label{tab:morphology_support}
\begin{adjustbox}{max width=\textwidth}
\begin{tabular}{lccrrrrrrr}
\toprule
\textbf{Parameter} & \textbf{Train min} & \textbf{Train max}
& \textbf{Go2} & \textbf{Go1} & \textbf{A1}
& \textbf{Cheetah} & \textbf{AlienGo} & \textbf{ANYmal} & \textbf{B1} \\
\midrule
Thigh mass
& 0.634 & 1.152
& 1.152 & 0.899 & 1.013
& 0.634 & 0.639 & 0.740 & \hi{4.300} \\

Calf mass
& 0.064 & 0.166
& 0.154 & 0.158 & 0.166
& 0.064 & \hi{0.207} & \hi{0.337} & \hi{1.050} \\

Hip mass
& 0.510 & 0.696
& 0.678 & 0.510 & 0.696
& 0.540 & \hi{1.000} & \hi{0.740} & \hi{2.150} \\

Foot mass
& 0.000 & 0.060
& 0.040 & 0.060 & 0.060
& 0.060 & 0.060 & \hi{0.250} & 0.050 \\

Base mass
& 3.300 & 6.921
& 6.921 & 4.800 & 6.000
& 3.300 & \hi{11.644} & \hi{18.210} & \hi{29.450} \\

Hip limit min
& -1.100 & -0.800
& -1.047 & -0.803 & -0.803
& \lo{-1.600} & \lo{-1.222} & \hi{-0.720} & \hi{-0.750} \\

Hip limit max
& 0.800 & 1.100
& 1.047 & 0.803 & 0.803
& \hi{1.600} & \hi{1.222} & \lo{0.720} & \lo{0.750} \\

Thigh limit min
& -1.650 & -1.050
& -1.571 & \hi{-1.047} & \hi{-1.047}
& \lo{-2.600} & \lo{-3.142} & -1.650 & \hi{-1.000} \\

Thigh limit max
& 3.490 & 4.050
& 3.491 & \hi{4.189} & \hi{4.189}
& \lo{2.600} & \lo{3.142} & 4.050 & 3.500 \\

Calf limit min
& -2.720 & -2.600
& \lo{-2.723} & -2.697 & -2.697
& -2.600 & \lo{-2.775} & -2.720 & -2.600 \\

Calf limit max
& -0.950 & -0.780
& -0.838 & -0.916 & -0.916
& -0.800 & \hi{-0.646} & -0.780 & \hi{-0.600} \\
\bottomrule
\end{tabular}
\end{adjustbox}
\end{table*}

\section{Morphology Descriptor and Randomization}

MorFiC uses an 11-D morphology vector containing link masses and joint-limit parameters. During training, each environment samples a morphology vector from the support in Table~\ref{tab:morphology_support}; this vector is applied in simulation and kept fixed for the episode. The same vector is encoded into the latent $z_m$ used by the actor and critic. Table~\ref{tab:morphology_support} also reports the descriptors of all evaluation robots, with out-of-support values marked by $\uparrow$ or $\downarrow$. Go1 and A1 remain largely near the training support, while AlienGo, ANYmal-C, and especially B1 introduce larger shifts in base mass, limb masses, and joint limits.

\subsection{Morphology Randomization}

The randomization range was chosen to cover moderate morphology variation around the Go2 source template while preserving stable PPO training. This creates a controlled zero-shot setting: MorFiC is not trained on the target robot models, but observes morphology variations that partially cover nearby quadrupeds. B1 is the most distant target, with substantially larger link and base masses than the training support.

\subsection{Effect of Expanding Morphology Randomization}

We also tested whether widening the morphology range improves transfer to distant robots such as B1. Expanding the support to include much heavier morphologies made the policy more conservative, reducing forward tracking without resolving the B1 failure mode (see Table~\ref{tab:film_morph_range_ablation}). This suggests that distant morphology transfer is not solved by naively widening randomization alone, and may require structured morphology curricula or limited multi-source training.

\begin{table}[t]
\centering
\small
\setlength{\tabcolsep}{3.0pt}
\renewcommand{\arraystretch}{1.05}
\caption{
\textbf{Effect of expanded morphology ranges and FiLM conditioning.}
Each entry reports $v_x^{\max}\pm\sigma / S$, where $S$ denotes survival rate.
Bold indicates the higher value for each metric independently; ties are bolded for both methods.
}
\label{tab:film_morph_range_ablation}
\begin{tabular}{lcc}
\toprule
\textbf{Target}
& \begin{tabular}{c}
\textbf{Expanded range} \\
$v_x^{\max}{\pm}\sigma / S$
\end{tabular}
& \begin{tabular}{c}
\textbf{FiLM} \\
$v_x^{\max}{\pm}\sigma / S$
\end{tabular} \\
\midrule

Go2
& $1.9013{\pm}0.3510 / 0.8531$
& $\mathbf{2.0415{\pm}0.3670} / \mathbf{0.9844}$ \\

Go1
& $1.9208{\pm}0.3087 / \mathbf{0.7656}$
& $\mathbf{2.0649{\pm}0.4552} / 0.5469$ \\

A1
& $1.8241{\pm}0.2322 / \mathbf{0.7596}$
& $\mathbf{2.1404{\pm}0.3608} / 0.5156$ \\

AlienGo
& $1.5499{\pm}0.5078 / \mathbf{1.0000}$
& $\mathbf{1.9812{\pm}0.5146} / \mathbf{1.0000}$ \\

B1
& $0.2001{\pm}0.0215 / \mathbf{1.0000}$
& $\mathbf{0.2109{\pm}0.0768} / \mathbf{1.0000}$ \\

Cheetah
& $1.2362{\pm}0.4805 / 0.3348$
& $\mathbf{1.3922{\pm}0.6902} / \mathbf{0.4219}$ \\

ANYmal-C
& $\mathbf{0.3079{\pm}0.1803} / \mathbf{1.0000}$
& $0.2505{\pm}0.2147 / 0.9688$ \\

\bottomrule
\end{tabular}
\end{table}




\section{Policy Architecture, Observation Streams, and PPO Hyperparameters}

\subsection{Observation Streams}

Table~\ref{tab:observations} summarizes the observation streams used by MorFiC. 
The policy receives a history of proprioceptive observations, including base velocity, angular velocity, projected gravity, command, joint positions, joint velocities, and previous actions. 
The privileged stream contains simulator-side physical parameters used during training, while the morphology stream contains the 11-D descriptor encoded into $z_m$. 
All methods use the same observation history and privileged information; they differ only in how morphology information is injected into the actor--critic architecture.

\subsection{Actor-Critic and Morphology Encoder Architecture}

Table~\ref{tab:architecture} \& ~\ref{tab:conditioning_variants} reports the architecture used for MorFiC. 
We use separate actor and critic MLP trunks with ELU activations. 
The morphology vector is encoded using a lightweight MLP and used to condition the critic through residual scaled FiLM modulation at each hidden layer. 
The final FiLM layer is initialized to zero, so training starts close to an unconditioned critic and gradually learns morphology-dependent modulation.

\subsection{PPO Hyperparameters}

Table~\ref{tab:ppo_hparams} lists the PPO hyperparameters used across all baseline and MorFiC experiments. 
All variants share the same rollout setup, learning rate schedule, clipping parameters, entropy coefficient, value loss coefficient, and optimization settings. 
Thus, performance differences are attributable to the morphology-conditioning mechanism rather than changes in PPO training.






\begin{table}[t]
\centering
\small
\setlength{\tabcolsep}{5pt}
\caption{Observation streams used for policy and value learning.}
\label{tab:observations}
\renewcommand{\arraystretch}{1.05}
\begin{tabular}{@{}l l r@{}}
\toprule
Stream & Signals & Dim \\
\midrule
$o_t$
& $[\,v,\ \omega,\ g_{\text{proj}},\ c,\ q-q_0,\ \dot q,\ a_{t-1}\,]$
& 48 \\
\midrule
$h_t$
& $\{o_{t-k}\}_{k=1}^{15}$
& $15\times48$ \\
\midrule
$p_t$
& $[\mu_{\text{fric}},\ e,\ m_{\text{base}},\ \Delta c,\ \mu_{\text{joint}}]$
& 18 \\
\midrule
$z_m$
& morphology encoder output
& $d=11$ \\
\bottomrule
\end{tabular}

\end{table}

\begin{table}[t]
\centering
\caption{Policy, critic, and morphology-conditioning architecture used in MorFiC. 
The PPO optimizer and rollout hyperparameters are shared across all methods and reported separately in Table~\ref{tab:ppo_hparams}.}
\label{tab:architecture}
\small
\begin{tabular}{ll}
\toprule
\textbf{Module} & \textbf{Specification} \\
\midrule
Actor MLP & $[\text{obs history}, \hat{z}_{\mathrm{priv}}, z_m] \rightarrow [512,256,128] \rightarrow a$, ELU \\
Critic trunk & $[\text{obs history}, z_{\mathrm{priv}}] \rightarrow [512,256,128] \rightarrow V$, ELU \\
Critic trunk (FiLM scaled) & $[\text{obs history}, z_{\mathrm{priv}}] \rightarrow [1024,512,256] \rightarrow V$, ELU \\
Initial action std. & $\sigma_0 = 1.0$ \\
Adaptation module & $\text{obs history} \rightarrow [256,128] \rightarrow \hat{z}_{\mathrm{priv}}$, ELU \\
Student latent dropout & $p=0.2$ during training \\
Morphology encoder & $d_m \rightarrow 128 \rightarrow 64$, ELU \\
Main critic conditioning & Residual scaled FiLM at each critic hidden layer \\
FiLM generator & $64 \rightarrow 128 \rightarrow 2H$, ELU \\
FiLM update & $h \leftarrow h(1+\gamma) + \beta$ \\
FiLM scale & $0.1$ \\
FiLM initialization & Last FiLM layer weights and biases initialized to zero \\
Decoder & Disabled \\
\bottomrule
\end{tabular}
\end{table}

\begin{table}[t]
\centering
\caption{Morphology-conditioning variants evaluated in the ablation study. 
All variants use the same PPO settings and base actor--critic widths as Table~\ref{tab:architecture}.}
\label{tab:conditioning_variants}
\small
\begin{tabular}{lll}
\toprule
\textbf{Variant} & \textbf{Critic conditioning} & \textbf{Key detail} \\
\midrule
Vanilla PPO & None & Critic does not receive morphology information \\
Concat-$z$ & Direct concatenation & $z_m$ is appended to the critic input \\
FiLM & Feature-wise modulation & $h \leftarrow h(1+\gamma) + \beta$ \\
Scaled FiLM & Residual FiLM with scale & FiLM outputs multiplied by $0.1$ \\
AdaLN & LayerNorm + FiLM & $h \leftarrow \mathrm{LN}(h)$ before FiLM modulation \\
HyperNet & Low-rank residual adapter & $h \leftarrow W h + B_z A_z h$, rank $16$, scale $0.1$ \\
\bottomrule
\end{tabular}
\end{table}

\begin{table}[t]
\centering
\caption{PPO hyperparameters used in all baseline and MorFiC experiments.}
\small
\setlength{\tabcolsep}{5pt}
\begin{tabular}{l c}
\hline
\textbf{PPO setting} & \textbf{Value} \\
\hline
Discount $\gamma$ & 0.99 \\
GAE $\lambda$ & 0.95 \\
Clip $\epsilon$ & 0.2 \\
Entropy coeff. & 0.01 \\
Value loss coeff. & 1.0 \\
Clipped value loss & True \\
Learning rate & $1{\times}10^{-3}$ \\
Schedule & adaptive (KL-targeted) \\
Target KL & 0.01 \\
Epochs / update & 5 \\
Mini-batches / update & 4 \\
Grad norm clip & 1.0 \\
\hline
\end{tabular}

\label{tab:ppo_hparams}
\end{table}

\section{Reward Terms and Weights}
We shape our MorFiC rewards as a weighted sum of (i) command tracking terms for linear and angular velocities using exponential, (ii) stability rewards that encourage upright posture and dynamic balance via a center-of-pressure (CoP) proximity reward, (iii) gait/contact shaping terms that match desired contact schedules by penalizing unwanted swing forces and stance-phase foot velocities and a detailed definition and weights are given in Table V.

\subsection{Reward structure}

Our reward structure is divided into three main categories: Task/Tracking, Penalties and stability shaping rewards, and the total reward sum is given by \eqref{eq:rews}.

\begin{equation}
\begin{aligned}
r_t \;=\;& \sum_{i \in T} w_i\, r_t^{(i)} \;-\; \sum_{j \in P} w_j\, p_t^{(j)} 
&+\; \sum_{k \in S} w_k\, s_t^{(k)} .
\end{aligned}
\label{eq:rews}
\end{equation}

 \textbf{{Task/Tracking}} rewards encourage the robot to complete the commanded locomotion objective. Specifically, we tracked the commanded linear velocity and the angular velocity using exponential rewards. (Table~\ref{tab:rewards} )
 
 \textbf{Penalties} are assigned as a way to discourage unsafe  movement, which could lead to collisions, bad gait, or physically collapse(like jerks, jerky motions, base collision, joint limit violation, etc). For example, we penalize body collisions
\begin{equation}
p_{\mathrm{col}} \;=\; \sum_{b \in B} {1.} \!\left(\left\lVert \mathbf{f}_{b} \right\rVert > f_{\min}\right),
\label{eq:collision_count}
\end{equation}

joint limit violations and is given by;
\begin{equation}
p_{\mathrm{lim}} \;=\; \sum_{i} \Big(
\max\!\left(0,\; q_i - q_i^{\max}\right)
+\max\!\left(0,\; q_i^{\min} - q_i\right)
\Big).
\label{eq:joint_limit_penalty}
\end{equation}

These help the robot in not deviating from the desired locomotion gait.

\textbf{Stability Shaping.} Motivated by our repeated failure on unseen morphologies, we introduce 3 lightweight stability rewards  namely upright stability, CoP stability, and feet-contact stability to reduce catastrophic tipping and improve recovery.

\begin{enumerate}
    \item \textbf{Upright stability.} Encourages recovery from large pitch/tilt by penalizing deviation from projected gravity.

\begin{equation}
s_{\mathrm{upright}}
= \exp\!\left(
-\frac{\left\lVert \mathbf{g}_{xy} \right\rVert^{2}}{\sin^{2}(\theta_{0})}
\right).
\label{eq:upright_stability}
\end{equation}

 \item \textbf{Center-of-pressure (CoP) stability.} Promotes balanced load distribution across the support polygon.

\begin{equation}
s_{\mathrm{CoP}}
= \exp\!\left(
-\frac{\left\lVert \mathrm{CoP}-\mathrm{center} \right\rVert}{\tau}
\right)\cdot {1}\!\left(n_{\mathrm{contact}}\ge 2\right).
\label{eq:cop_stability}
\end{equation}

\item  \textbf{Feet-contact / stance-width stability.} Penalize  overly narrow instances during unstable locomotion (caused by large tilt or rapid movement when unstable). We  gate this using eqns. \ref{eq:width_gate}, \ref{eq:stance_width}. 

\begin{equation}
\mathrm{gate}
= 1\!\left(
\left\lVert \mathbf{g}_{xy} \right\rVert^{2}>\eta \;\vee\;
\left\lVert \mathbf{v}_{xy} \right\rVert>v_{\mathrm{th}}
\right),
\label{eq:width_gate}
\end{equation}

and define 
\begin{equation}
s_{\mathrm{width}}
= \mathrm{gate}\cdot
\exp\!\left(
-\frac{\max(0,\, w_{\min}-w)}{\sigma_{w}}
\right)
+ (1-\mathrm{gate}).
\label{eq:stance_width}
\end{equation}

\end{enumerate}

In addition to these components, we include a small set of rewards adapted from earlier work \cite{margolis2023walk}. \textbf{All baselines were trained with the same reward}, so as to clearly isolate the performance difference caused by the poor approximation of the value function.

\begin{table}[t]
\centering
\caption{\textbf{Reward terms} used in our MorFiC locomotion objective. $w_k$ are set in the reward configuration; fixed scalings appearing in code (e.g., the factor 4 in $r_{\text{lin}}$) are shown explicitly.}
\vspace{1.00em}
\small
\setlength{\tabcolsep}{4pt}
\begin{tabular}{l l l}
\hline
\textbf{Term} & \textbf{Definition} & \textbf{Weight} \\
\hline
$r_{\text{lin}}$ &
$4\exp\!\left(-\frac{\|\mathbf{v}^{cmd}_{xy}-\mathbf{v}_{xy}\|^2}{\sigma_{\text{lin}}}\right)$
& $w_{\text{lin}}$ \\

$r_{\text{yaw}}$ &
$\exp\!\left(-\frac{(\omega^{cmd}_z-\omega_z)^2}{\sigma_{\text{yaw}}}\right)$
& $w_{\text{yaw}}$ \\

$p_{v_z}$ & $(v_z)^2$ & $w_{v_z}$ \\
$p_{\omega_{xy}}$ & $\|\boldsymbol{\omega}_{xy}\|^2$ & $w_{\omega}$ \\
$p_{\text{ori}}$ & $\|\hat{\mathbf{g}}_{xy}\|^2$ & $w_{\text{ori}}$ \\

$p_{\tau}$ & $\|\boldsymbol{\tau}\|^2$ & $w_{\tau}$ \\
$p_{\ddot{q}}$ & $\left\|\frac{\dot{q}_{t-1}-\dot{q}_t}{\Delta t}\right\|^2$ & $w_{\ddot{q}}$ \\
$p_{\Delta a}$ & $\|a_{t-1}-a_t\|^2$ & $w_{\Delta a}$ \\

$p_{\text{coll}}$ &
$\sum_{b\in\mathcal{B}} \mathbb{I}\!\left(\|\mathbf{f}_b\|>0.1\right)$
& $w_{\text{coll}}$ \\

$p_{\text{jl}}$ &
$\sum_i \left[(q_i-q_i^{\min})_{-} + (q_i-q_i^{\max})_{+}\right]$
& $w_{\text{jl}}$ \\

$r_{\text{upright}}$ &
$\exp\!\left(-\frac{\|\hat{\mathbf{g}}_{xy}\|^2}{\sin^2(\theta_0)}\right),\;\theta_0{=}0.3$
& $w_{\text{upright}}$ \\

$r_{\text{cop}}$ &
$\mathbb{I}(n_c\ge 2)\exp\!\left(-\frac{\|\text{CoP}-\text{center}\|}{\tau}\right),\;\tau{=}0.10$
& $w_{\text{cop}}$ \\

$r_{\text{width}}$ &
\emph{(yaw-frame)} $\exp\!\left(-\frac{(w_{\min}-w)_{+}}{\sigma_w}\right)$ (gated)
& $w_{\text{width}}$ \\

$r_{\text{gaitF}}$ &
$-\frac{1}{4}\sum_i (1-d_i)\left(1-\exp(-f_i^2/\sigma_F)\right)$
& $w_{\text{gaitF}}$ \\

$r_{\text{gaitV}}$ &
$-\frac{1}{4}\sum_i d_i\left(1-\exp(-\|\dot{\mathbf{x}}_i\|^2/\sigma_V)\right)$
& $w_{\text{gaitV}}$ \\

$p_{\text{slip}}$ &
$\sum_i \mathbb{I}(c_i)\|\dot{\mathbf{x}}_{i,xy}\|^2$
& $w_{\text{slip}}$ \\

$p_{\text{impact}}$ &
$\sum_i \mathbb{I}(c_i)\,\text{clip}(\dot{z}_{i,t-1},-100,0)^2$
& $w_{\text{impact}}$ \\

$p_{f}$ &
$\sum_i \big(\|\mathbf{f}_i\| - f_{\max}\big)_{+}$
& $w_{f}$ \\

$p_{q}$ & $\|q-q^{0}\|^2$ & $w_{q}$ \\
$p_{\dot{q}}$ & $\|\dot{q}\|^2$ & $w_{\dot{q}}$ \\

$r_{\text{jump}}$ &
$-(z - (h^{cmd}+h_0))^2$
& $w_{\text{jump}}$ \\

$r_{\text{raibert}}$ &
$\|\mathbf{x}^{des}_{foot}-\mathbf{x}_{foot}\|^2$ (body yaw frame)
& $w_{\text{raibert}}$ \\
\hline
\end{tabular}

\label{tab:rewards}
\end{table}

\section{Training Details}

\subsection{Random Seeds and Reporting}
Unless stated otherwise, each internal PPO and MorFiC variant is trained with $n=3$ independent random seeds. For simulation evaluation, each checkpoint is evaluated using the same target robots, command settings, rollout horizon, and number of parallel environments. Reported values are averaged across evaluation rollouts; standard deviations are reported over rollout-level measurements. For external baselines, we use the available released checkpoints or reproduced checkpoints as described in the corresponding baseline setup.

\subsection{Simulation and Training Setup}
All policies are trained in IsaacGym Simulator~\cite{makoviychuk2021isaac} with 4096 parallel environments with a time-step of $dt$ = 0.005s; episode lasts roughly 20s. A command curriculum and domain randomization (joint friction, motor delays, sensor noise) are applied identically across all baselines. We used the same PPO hyperparameters, reward functions, command curriculum, morphology ranges, and network sizes across all baselines variants (See Appendix). And finally, each policy is trained a total of 400 million time steps utilizing a Nvidia RTX 4090 laptop-based GPU, requiring roughly 2 hours of wall clock time. We have adapted our code mainly from open-source repositories~\cite{margolis2023walk,rudin2022learning}.

\subsection{Domain Randomization}
We apply domain randomization during training to improve robustness and sim-to-real transfer. Randomization is applied periodically with the interval $10$\,s (rigid-body properties may be randomized after initialization). We randomize the properties of contact with the ground by sampling the friction coefficient uniformly from $[0.5,\,1.25]$. We inject external disturbances by pushing the robot at fixed intervals of $15$\,s with a maximum planar push velocity of $1.0$\,m/s. To model actuation/communication delay, we randomize action latency using a history buffer of up to 6 timesteps (randomized lag timesteps enabled).

\subsection{Evaluation Fairness}

 In order to maintain the fairness of the evaluation, we kept all the necessary parameters like morphological and domain randomization parameters for the training robot, including all the PPO hyper-parameters, command curriculum ranges and reward functions, latent dimensions the same across all the variants.

\section{Additional Transfer Results}

\subsection{Additional Results MorFiC Vs. external baselines}

Table~\ref{tab:cross_embodiment_baselines_25} reports an additional evaluation at a higher commanded forward speed, $v_x^{cmd}=2.5$\,m/s. This setting tests whether the transfer trends from the main $2.0$\,m/s evaluation persist under a more aggressive command. MorFiC-F remains competitive on the source and nearby morphologies, while performance degrades on more distant targets such as B1 and ANYmal-C. As in the main results, survival is interpreted jointly with achieved forward velocity, since high survival with negligible motion does not indicate successful transfer.

\begin{table}[t]
\centering
\small
\setlength{\tabcolsep}{1.0pt}
\renewcommand{\arraystretch}{1.05}
\caption{
\textbf{External baselines vs. MorFiC at higher commanded speed.}
Each entry reports $v_x^{\max}\pm\sigma / S$ (m/s; survival).
All methods are evaluated at $v_x^{\mathrm{cmd}}=2.5$ m/s, except GenLoco, which does not take a commanded forward velocity as input.
Bold indicates the best value for each metric independently.
}
\label{tab:cross_embodiment_baselines_25}
\resizebox{\columnwidth}{!}{
\begin{tabular}{lccccc}
\toprule
\textbf{Target}
& \begin{tabular}{c}\textbf{URMA}\\[-1pt] $v_x^{\max}{\pm}\sigma/S$\end{tabular}
& \begin{tabular}{c}\textbf{BodyTrans.}\\[-1pt] $v_x^{\max}{\pm}\sigma/S$\end{tabular}
& \begin{tabular}{c}\textbf{GenLoco}\\[-1pt] $v_x^{\max}{\pm}\sigma/S$\end{tabular}
& \begin{tabular}{c}\textbf{Embod.}\\[-1pt] $v_x^{\max}{\pm}\sigma/S$\end{tabular}
& \begin{tabular}{c}\textbf{MorFiC-F}\\[-1pt] $v_x^{\max}{\pm}\sigma/S$\end{tabular} \\
\midrule

Go2
& $2.389{\pm}0.405/\mathbf{1.000}$
& $0.313{\pm}0.232/0.625$
& $0.592{\pm}0.047/0.922$
& $1.072{\pm}0.092/0.859$
& $\mathbf{2.405{\pm}0.405}/0.859$ \\

Go1
& $\mathbf{2.514{\pm}0.430}/\mathbf{1.000}$
& $0.206{\pm}0.151/\mathbf{1.000}$
& $0.985{\pm}0.007/\mathbf{1.000}$
& $2.343{\pm}0.131/0.797$
& $1.972{\pm}0.442/0.516$ \\

A1
& $\mathbf{2.230{\pm}0.336}/\mathbf{1.000}$
& $0.374{\pm}0.183/0.515$
& $0.919{\pm}0.011/\mathbf{1.000}$
& $2.143{\pm}0.116/0.719$
& $1.595{\pm}0.467/0.156$ \\

AlienGo
& $-/0.000$
& $0.358{\pm}0.315/0.203$
& $\mathbf{0.964{\pm}0.032}/\mathbf{1.000}$
& $0.452{\pm}0.085/0.000$
& $0.832{\pm}0.542/\mathbf{1.000}$ \\

B1
& $-/0.000$
& $0.080{\pm}0.140/0.640$
& $0.00/0.000$
& $\mathbf{3.062{\pm}0.214}/\mathbf{0.797}$
& $0.121{\pm}0.060/0.000$ \\

Cheetah
& $-/0.000$
& $-/\mathbf{1.000}$
& $0.729{\pm}0.061/0.969$
& $-0.118{\pm}0.091/0.000$
& $\mathbf{1.128{\pm}0.986}/0.234$ \\

ANYmal-C
& $\mathbf{2.159{\pm}0.354}/\mathbf{1.000}$
& $0.014{\pm}0.081/0.843$
& $0.708{\pm}0.005/\mathbf{1.000}$
& $0.359{\pm}0.150/0.453$
& $0.303{\pm}0.422/0.984$ \\

\bottomrule
\end{tabular}
}
\end{table}



\begin{table}[t]
\centering
\scriptsize
\setlength{\tabcolsep}{1.0pt}
\renewcommand{\arraystretch}{1.05}
\caption{
\textbf{Extended critic calibration diagnostics across PPO variants.}
Each entry reports $|\mathrm{NVB}|$/NVE/ASA, where NVB is the normalized value bias,
NVE is the normalized value error, and ASA is the advantage-sign agreement.
Lower $|\mathrm{NVB}|$ and NVE are better, while higher ASA is better.
Best value for each metric within a target row is bolded.
}
\label{tab:critic_metrics_variants}
\resizebox{\columnwidth}{!}{
\begin{tabular}{lcccc}
\toprule
\textbf{Target}
& \begin{tabular}{c}\textbf{Vanilla PPO}\\[-1pt] $|\mathrm{NVB}|$/NVE/ASA\end{tabular}
& \begin{tabular}{c}\textbf{Concat-$z$}\\[-1pt] Actor only\end{tabular}
& \begin{tabular}{c}\textbf{Concat-$z$}\\[-1pt] Actor+Critic\end{tabular}
& \begin{tabular}{c}\textbf{FiLM}\\[-1pt] $|\mathrm{NVB}|$/NVE/ASA\end{tabular} \\
\midrule

Go2
& $1.944/1.981/0.610$
& $\mathbf{0.177}/1.168/0.689$
& $0.535/\mathbf{0.960}/{0.670}$
& $1.768/1.943/\textbf{0.771}$ \\

Go1
& $1.591/1.817/0.612$
& $\mathbf{0.068}/\mathbf{1.096}/0.662$
& $1.623/1.834/0.674$
& $1.375/1.653/\mathbf{0.681}$ \\

A1
& $2.027/2.203/0.603$
& $2.033/16.463/\mathbf{0.754}$
& $\mathbf{0.199}/\mathbf{0.936}/0.701$
& $1.482/1.813/0.675$ \\

AlienGo
& ${0.070}/1.454/0.612$
& $0.468/1.209/0.569$
& $0.153/1.117/0.607$
& $\textbf{0.048}/\mathbf{0.835}/\mathbf{0.706}$ \\

B1
& $0.182/1.231/0.618$
& $0.358/1.187/0.570$
& $\mathbf{0.040}/{1.134}/\mathbf{0.624}$
& $0.645/\textbf{1.279}/0.618$ \\

Cheetah
& $0.362/1.869/0.428$
& $\mathbf{0.220}/1.086/0.638$
& $1.085/1.344/0.730$
& $0.464/\mathbf{0.947}/\mathbf{0.768}$ \\

ANYmal-C
& $\mathbf{0.091}/1.229/0.514$
& $0.411/\mathbf{1.227}/0.528$
& $0.385/1.307/\mathbf{0.620}$
& $0.120/1.229/0.578$ \\

\bottomrule
\end{tabular}
}
\end{table}

\begin{table}[t]
\centering
\scriptsize
\setlength{\tabcolsep}{1.0pt}
\renewcommand{\arraystretch}{1.05}
\caption{
\textbf{Extended critic calibration diagnostics for external baselines.}
Each entry reports $|\mathrm{NVB}|$/NVE/ASA, where NVB is normalized value bias,
NVE is normalized value error, and ASA is advantage-sign agreement.
Lower $|\mathrm{NVB}|$ and NVE are better, while higher ASA is better.
Best value for each metric within a target row is bolded.
}
\label{tab:critic_metrics_external}
\resizebox{\columnwidth}{!}{
\begin{tabular}{lccccc}
\toprule
\textbf{Target}
& \begin{tabular}{c}\textbf{Embod. Scale}\\[-1pt] $|\mathrm{NVB}|$/NVE/ASA\end{tabular}
& \begin{tabular}{c}\textbf{URMA}\\[-1pt] $|\mathrm{NVB}|$/NVE/ASA\end{tabular}
& \begin{tabular}{c}\textbf{BodyTrans.}\\[-1pt] $|\mathrm{NVB}|$/NVE/ASA\end{tabular}
& \begin{tabular}{c}\textbf{GenLoco}\\[-1pt] $|\mathrm{NVB}|$/NVE/ASA\end{tabular}
& \begin{tabular}{c}\textbf{MorFiC-F}\\[-1pt] $|\mathrm{NVB}|$/NVE/ASA\end{tabular} \\
\midrule

Go2
& $3.437/3.646/0.657$
& $3.005/3.178/\mathbf{0.702}$
& $3.319/3.492/0.652$
& $5.458/5.674/0.561$
& $\mathbf{1.768}/\mathbf{1.943}/0.671$ \\

Go1
& $2.569/2.772/0.607$
& $3.170/3.333/0.692$
& $4.771/4.983/\mathbf{0.724}$
& $3.402/3.594/0.623$
& $\mathbf{1.375}/\mathbf{1.653}/0.681$ \\

A1
& $2.368/2.585/0.599$
& $3.390/3.551/\mathbf{0.686}$
& $3.853/4.032/0.686$
& $2.956/3.128/0.612$
& $\mathbf{1.482}/\mathbf{1.813}/0.675$ \\

AlienGo
& -- 
& -- 
& $3.104/3.476/0.661$
& $1.170/1.522/\mathbf{0.709}$
& $\mathbf{0.098}/\mathbf{0.835}/0.706$ \\

B1
& $3.330/3.590/0.657$
& --
& $3.571/3.781/\mathbf{0.694}$
& $7.428/7.729/0.667$
& $\mathbf{0.645}/\mathbf{1.279}/0.618$ \\

Cheetah
& --
& --
& --
& $5.294/5.511/0.594$
& $\mathbf{0.464}/\mathbf{0.947}/\mathbf{0.768}$ \\

ANYmal-C
& $3.589/3.812/0.667$
& $2.538/2.740/0.728$
& $3.368/3.529/0.674$
& $1.107/2.324/\mathbf{0.733}$
& $\mathbf{0.120}/\mathbf{1.229}/0.578$ \\

\bottomrule
\end{tabular}
}
\end{table}

\section{Extended Critic Calibration Diagnostics}

The main paper reports explained variance as the primary critic-calibration metric. Here, we provide additional diagnostics that capture complementary aspects of value-function quality. Specifically, we report $|\mathrm{NVB}|$, NVE, and ASA for both internal PPO variants and external baselines. Lower $|\mathrm{NVB}|$ indicates smaller systematic value bias, lower NVE indicates smaller value prediction error, and higher ASA indicates better agreement between critic-induced and Monte-Carlo advantage signs. 

Table~\ref{tab:critic_metrics_variants} compares the internal PPO variants. The results show that additive morphology conditioning does not consistently resolve critic error across morphologies. FiLM improves the strongest OOD cases such as AlienGo and Cheetah, where it achieves lower NVE and higher ASA than the vanilla and concatenation-based variants. This supports the claim that multiplicative critic conditioning improves the quality of the value baseline under morphology shift. 

Table~\ref{tab:critic_metrics_external} reports the same diagnostics for external baselines. Although some external methods achieve reasonable locomotion performance on certain robots, their value predictions often remain poorly calibrated, with large $|\mathrm{NVB}|$ and NVE. MorFiC-F generally gives substantially lower value error, especially on OOD targets such as AlienGo, B1, Cheetah, and ANYmal-C. ASA is more mixed, so we interpret it jointly with $|\mathrm{NVB}|$ and NVE rather than as a standalone measure.

\section{Fixed-Rollout Critic Causality Analysis}

To test whether morphology conditioning improves the actor through better critic estimates, we run a fixed-rollout diagnostic. For each trained policy, we freeze the sampled trajectories and compare the critic-induced advantage signs against an MC-return reference. We report the sign-flip rate, where lower values indicate fewer wrong-sign policy updates, and the gradient cosine against the MC-reference actor gradient, where higher values indicate better update alignment.

MorFiC generally yields lower flip rates and higher gradient alignment on near-support targets such as Go2, A1, AlienGo, and Cheetah, suggesting that FiLM conditioning improves the critic signal used by PPO. However, on distant morphologies such as B1 and ANYmal-C, the cosine can become negative, exposing a failure mode where the learned critic induces updates that are misaligned with the MC reference. This supports our diagnosis that cross-embodiment failure is not only an actor-transfer problem, but also a critic-miscalibration problem. (see Table~\ref{tab:causal_adv_all})

\begin{table*}[t]
\centering
\scriptsize
\setlength{\tabcolsep}{2.4pt}
\renewcommand{\arraystretch}{1.05}
\caption{
\textbf{Fixed-rollout critic diagnostic across target morphologies.}
Each entry reports Flip/Cos., where Flip is the sign-flip rate against the MC reference advantage
(lower is better) and Cos. is the actor-gradient cosine with the MC-reference gradient
(higher is better). Bold indicates the best learned method for each metric independently;
the MC row is the reference and is not considered for bolding.
}
\label{tab:causal_adv_all}
\resizebox{\textwidth}{!}{
\begin{tabular}{lcccccccccccccc}
\toprule
\textbf{Src.}
& \multicolumn{2}{c}{\textbf{Go2}}
& \multicolumn{2}{c}{\textbf{Go1}}
& \multicolumn{2}{c}{\textbf{A1}}
& \multicolumn{2}{c}{\textbf{AlienGo}}
& \multicolumn{2}{c}{\textbf{B1}}
& \multicolumn{2}{c}{\textbf{Cheetah}}
& \multicolumn{2}{c}{\textbf{ANYmal-C}} \\
\cmidrule(lr){2-3}
\cmidrule(lr){4-5}
\cmidrule(lr){6-7}
\cmidrule(lr){8-9}
\cmidrule(lr){10-11}
\cmidrule(lr){12-13}
\cmidrule(lr){14-15}
& Flip & Cos.
& Flip & Cos.
& Flip & Cos.
& Flip & Cos.
& Flip & Cos.
& Flip & Cos.
& Flip & Cos. \\
\midrule

MC
& 0.000 & 1.000
& 0.000 & 1.000
& 0.000 & 1.000
& 0.000 & 1.000
& 0.000 & 1.000
& 0.000 & 1.000
& 0.000 & 1.000 \\

MorFiC
& \textbf{0.227} & 0.880
& 0.187 & \textbf{0.972}
& \textbf{0.239} & \textbf{0.785}
& \textbf{0.255} & \textbf{0.366}
& 0.272 & -0.182
& \textbf{0.209} & \textbf{0.679}
& \textbf{0.402} & -0.149 \\

Concat
& 0.307 & 0.720
& \textbf{0.151} & 0.830
& 0.344 & 0.427
& 0.329 & 0.004
& 0.269 & 0.106
& 0.236 & 0.575
& 0.479 & -0.419 \\

Act-only
& 0.524 & \textbf{0.890}
& 0.502 & 0.924
& 0.511 & 0.717
& 0.318 & 0.342
& 0.565 & \textbf{0.647}
& 0.265 & 0.644
& 0.470 & \textbf{-0.044} \\

Vanilla PPO
& 0.235 & 0.865
& 0.229 & 0.684
& 0.260 & 0.653
& 0.322 & -0.113
& \textbf{0.176} & 0.601
& 0.253 & 0.264
& 0.461 & -0.074 \\

\bottomrule
\end{tabular}
}
\end{table*}

\section{Training Compute Methodology}

We report transition-normalized GPU-hours because the baselines use different training budgets. For each method, we profile the released training code on our local hardware for 100 iterations, measure the average wall-clock time per iteration, and compute the transitions per iteration from the method configuration. We then extrapolate total GPU-hours to the reported transition budget:
\[
\mathrm{GPU\text{-}hours}
=
\frac{T}{N_{\mathrm{env}}H}
\cdot
\frac{\tau_{\mathrm{iter}}}{3600},
\]
where $T$ is the total transition budget, $N_{\mathrm{env}}$ is the number of parallel environments, $H$ is the rollout horizon, and $\tau_{\mathrm{iter}}$ is the measured time per iteration. We also report GPU-hours normalized to 400M transitions for fair per-transition comparison. For BodyTransformer, we reproduce training to approximately the MorFiC budget; for larger-budget methods, we extrapolate using their reported transition counts.

\begin{table}[t]
\centering
\small
\setlength{\tabcolsep}{2.0pt}
\renewcommand{\arraystretch}{1.08}
\caption{
\textbf{Training compute comparison.}
We report total transitions, approximate PPO iterations, wall-clock seconds per iteration,
total single-GPU-equivalent training hours, and GPU-hours normalized to a 400M-transition budget.
Lower GPU-hours@400M indicates lower compute cost per transition.
Expert-pool methods are separated because they train many specialized policies rather than one generalist policy.
}
\label{tab:training_compute}
\resizebox{\columnwidth}{!}{
\begin{tabular}{lccccc l}
\toprule
\textbf{Method}
& \textbf{Transitions}
& \textbf{Iters.}
& \textbf{Sec/iter}
& \textbf{GPU-h}
& \begin{tabular}{c}\textbf{GPU-h}\\[-1pt]\textbf{@400M}\end{tabular}
& \textbf{Notes} \\
\midrule

MorFiC
& 400M
& 4,000
& \textbf{1.98--2.70}
& \textbf{2.2--3.0}
& \textbf{2.2--3.0}
& Single generalist policy; local run \\

BodyTransformer
& 400M
& 8,000
& 17.25
& 38.3
& 39.0
& Reproduced \texttt{model\_8000} \\

URMA
& 1.600B
& 3,064
& 366--392
& 311--333
& 77.8--83.3
& Original training budget \\

GenLoco
& 200M
& 1,526
& 125.8
& 53.3
& 106.6
& 200M setting \\

GenLoco
& 800M
& 6,104
& 125.8
& 213.5
& 106.8
& 800M setting \\

\midrule
\multicolumn{7}{l}{\textit{Expert-pool / selected-expert methods}} \\

Embod. Scaling, Gendog8 expert
& 1.720B
& 17,499
& 0.94
& 4.5
& 1.05
& One selected released expert \\

Embod. Scaling, full Gendog pool
& 570.6B
& $\sim$5.80M
& 0.94
& 1507
& 1.06
& 332 experts; $\sim$5 TB storage \\

\bottomrule
\end{tabular}
}
\end{table}



\section{Real-Robot Rollout Diagnostics}

\begin{figure}[t]
\centering
\includegraphics[width=\columnwidth]{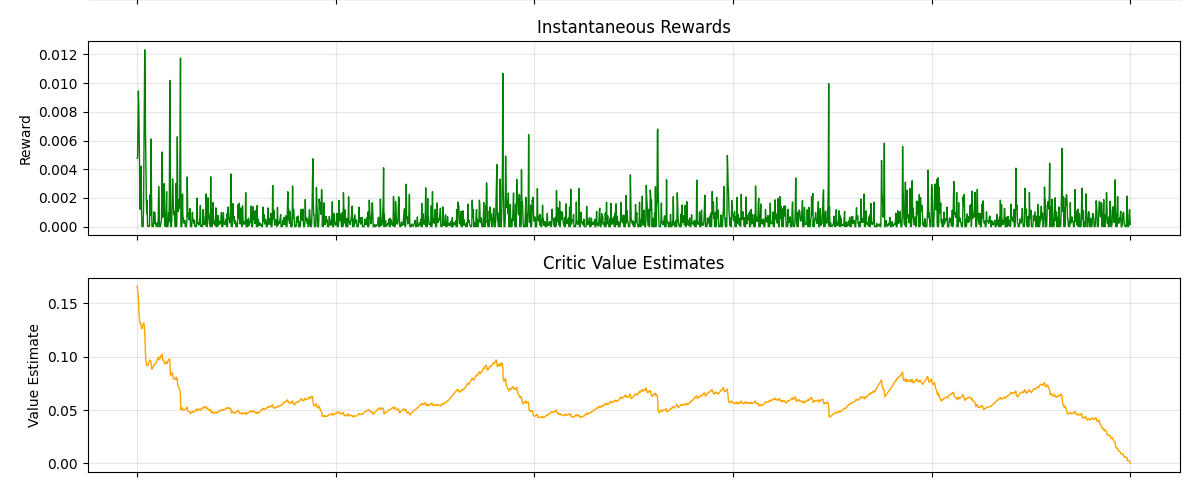}
\includegraphics[width=\columnwidth]{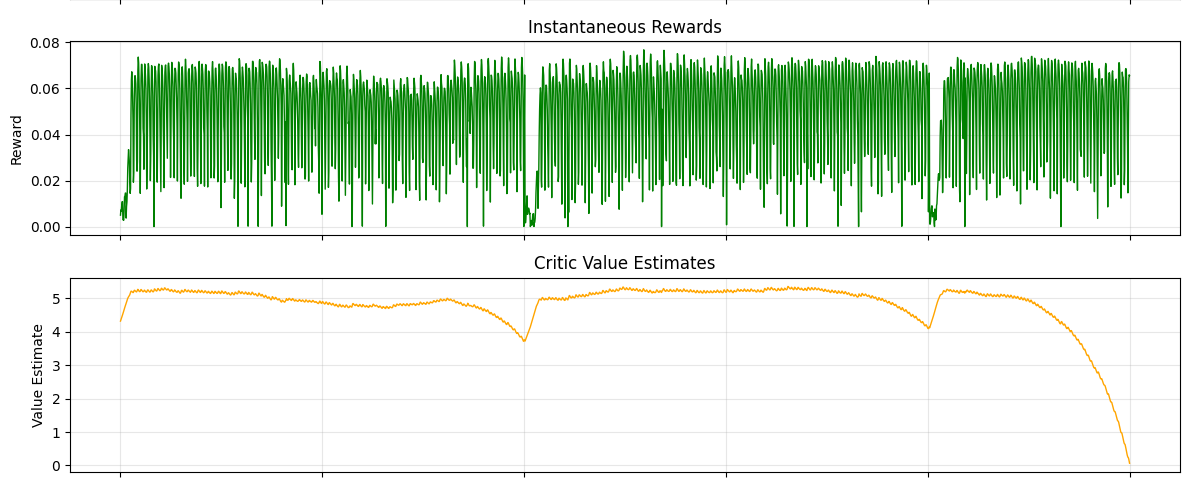}

\includegraphics[width=0.3\columnwidth]{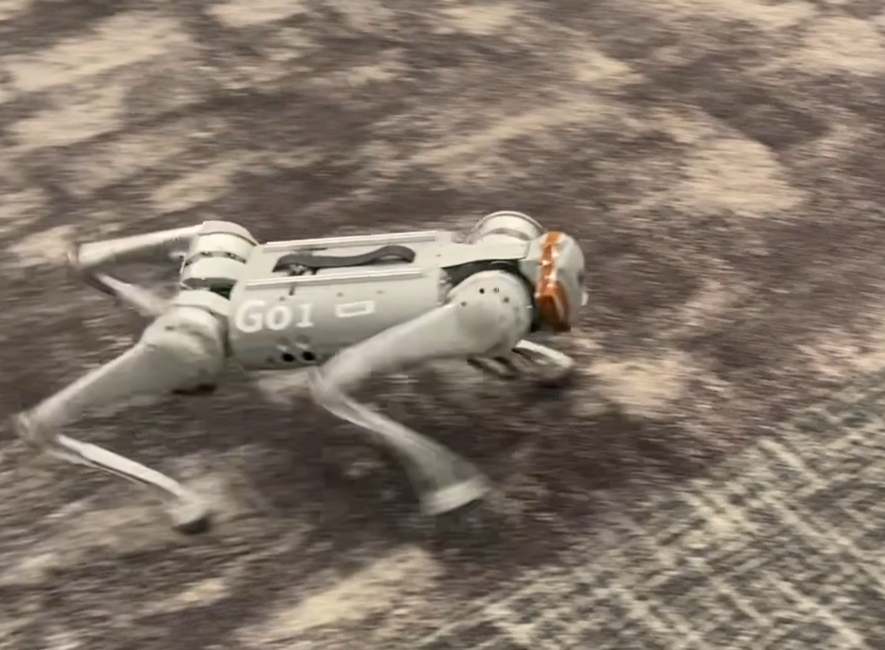}\hfill
\includegraphics[width=0.3\columnwidth]{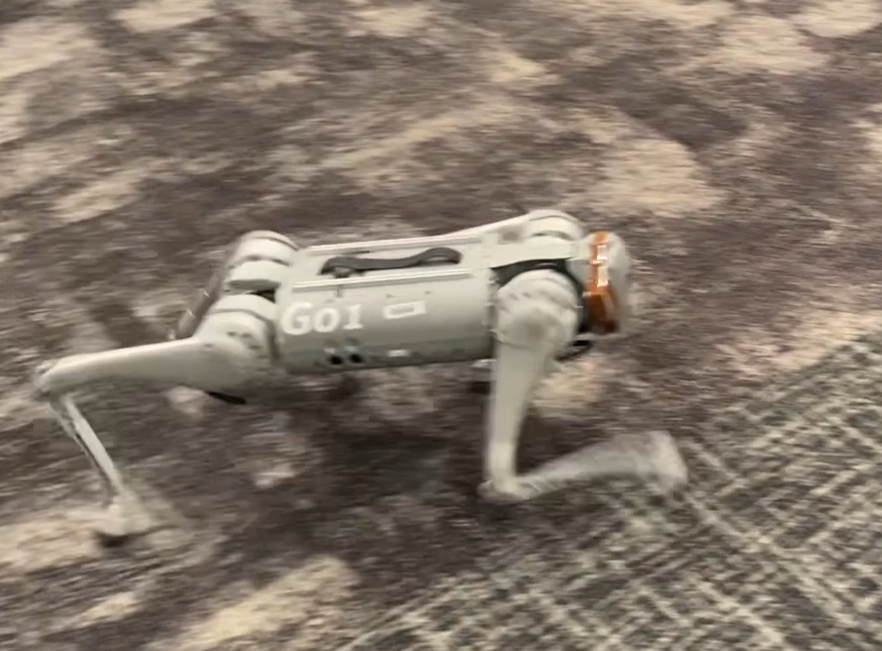}\hfill
\includegraphics[width=0.3\columnwidth]{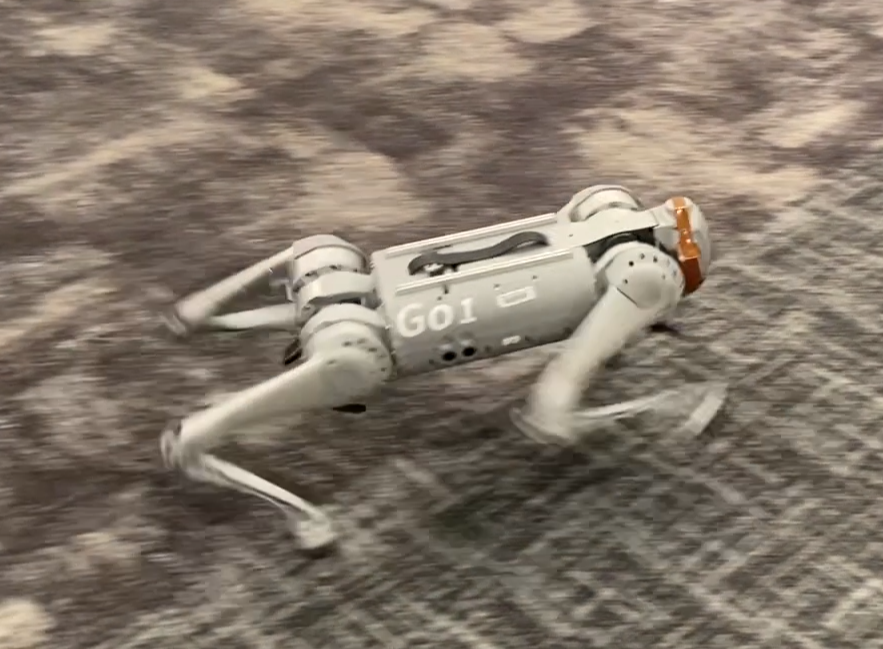}
\vspace{-1mm}
\caption{ Above figure shows representative reward and value trajectories during deployment on the Go1 robot. The MorFiC  produces smooth value estimates (orange) (second row) that track the expected discounted returns from noisy instantaneous rewards (green). While non-MorFic value estimates are noisy (first row), MorFiC's morphology-conditioned critic maintains stable value predictions even on unseen morphologies.}
\label{fig:critic}
\end{figure}

To examine whether the learned critic remains meaningful during real deployment, we record reward and value trajectories from Go1 hardware rollouts. Figure~\ref{fig:critic} shows that MorFiC produces smoother value estimates that follow the trend of the discounted return despite noisy instantaneous rewards. In contrast, the non-MorFiC critic exhibits higher variance and less stable predictions. This suggests that morphology-conditioned value estimation can improve critic stability beyond simulation and remains useful under real-world rollout noise.

\end{document}